\documentclass[10pt,twocolumn,letterpaper]{article}

\usepackage[caption = false]{subfig}
\usepackage{iccv}
\usepackage{times}
\usepackage{epsfig}
\usepackage{graphicx}
\usepackage{amsmath}
\usepackage{amssymb}
\usepackage{cite}
\usepackage{tabularx}
\usepackage{makecell}
\usepackage{multirow}
\usepackage{booktabs}

\usepackage[pagebackref=true,breaklinks=true,letterpaper=true,colorlinks,bookmarks=false]{hyperref}

\iccvfinalcopy 

\def\iccvPaperID{6085} 

\ificcvfinal\fi

\begin{document}
\newcolumntype{L}[1]{>{\raggedright\arraybackslash}p{#1}}
\newcolumntype{C}[1]{>{\centering\arraybackslash}p{#1}}
\newcolumntype{R}[1]{>{\raggedleft\arraybackslash}p{#1}}

\title{Asymmetric Bilateral Motion Estimation for Video Frame Interpolation}

\author{Junheum Park\\
Korea University\\
{\tt\small jhpark@mcl.korea.ac.kr}
\and
Chul Lee\\
Dongguk University\\
{\tt\small chullee@dongguk.edu}
\and
Chang-Su Kim\\
Korea University\\
{\tt\small changsukim@korea.ac.kr}
}

\maketitle
\ificcvfinal\thispagestyle{empty}\fi

\begin{abstract}
   We propose a novel video frame interpolation algorithm based on asymmetric bilateral motion estimation (ABME), which synthesizes an intermediate frame between two input frames. First, we predict symmetric bilateral motion fields to interpolate an anchor frame. Second, we estimate asymmetric bilateral motions fields from the anchor frame to the input frames. Third, we use the asymmetric fields to warp the input frames backward and reconstruct the intermediate frame. Last, to refine the intermediate frame, we develop a new synthesis network that generates a set of dynamic filters and a residual frame using local and global information. Experimental results show that the proposed algorithm achieves excellent performance on various datasets. The source codes and pretrained models are available at \url{https://github.com/JunHeum/ABME}.
\end{abstract}

\section{Introduction}

Video frame interpolation is a low-level vision task to increase the frame rate of a video sequence by interpolating intermediate frames between successive input frames. It is widely used in applications, including video enhancement \cite{xue2019toflow}, video compression \cite{Lu2018compression}, slow-motion generation \cite{jiang2018slomo}, and view synthesis \cite{Flynn2016viewsynthesis, kalantari2016viewsynthesis}. Due to its practical importance, various algorithms have been proposed to increase video frame rates \cite{choi2007motion, jeong2013texture, bao2018memc, bao2019dain, jiang2018slomo, liu2017dvf, liu2019cyclicgen, long2016learning, niklaus2017adaconv, niklaus2017sepconv, niklaus2018ctx, gui2020feflow, cheng2020dsepconv, choi2020cain, lee2020adacof, park2020bmbc, niklaus2020softsplatting}.

These algorithms can be classified into three categories: kernel-based \cite{niklaus2017adaconv, niklaus2017sepconv, bao2018memc, bao2019dain, cheng2020dsepconv, lee2020adacof}, phase-based \cite{meyer2015phase, meyer2018phase}, and motion-based \cite{liu2017dvf, liu2019cyclicgen, bao2018memc, bao2019dain, jiang2018slomo, niklaus2018ctx, gui2020feflow, park2020bmbc, niklaus2020softsplatting}. With the recent advances in optical flow estimation \cite{dosovitskiy2015flownet, ilg2017flownet2, ranjan2017spynet, sun2018pwc, zhao2020maskflownet, jonschkowski2020uflow, liu2020arflow}, motion-based algorithms have been developed most actively. They use optical flows to predict an intermediate frame by warping two successive frames forward or backward. For example, Niklaus and Liu \cite{niklaus2018ctx} predict bi-directional optical flows between two frames and halve them to generate intermediate frames based on forward warping. However, forward warping may cause interpolation artifacts in holes and overlapped regions \cite{wolberg2000warping}. To overcome the hole issue, they develop a synthesis network that learns to fill in holes. The overlapping issue, however, remains, so they propose softmax-splatting \cite{niklaus2020softsplatting} to combine overlapping pixel information adaptively and render the target pixel more faithfully.

\begin{figure}
  \centering
  \subfloat[] {\includegraphics[width=0.7\linewidth]{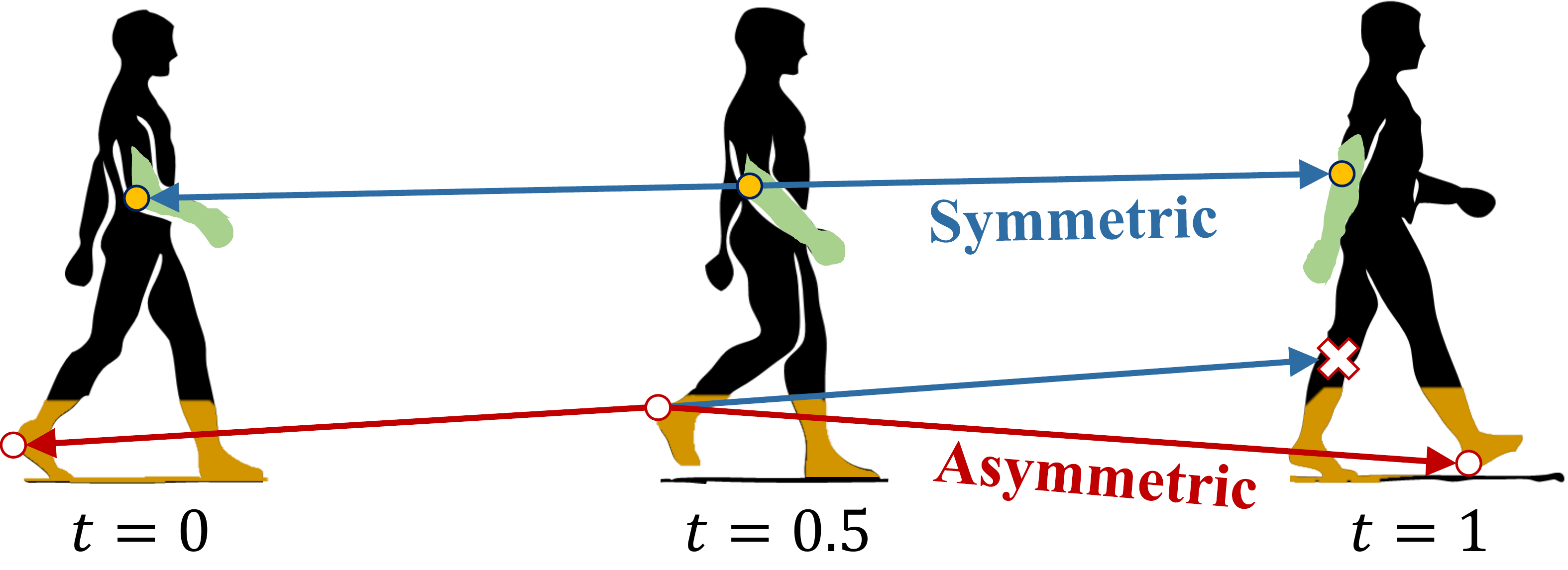}\label{fig:toy_example}}\\
  \subfloat[]{\includegraphics[width=\linewidth]{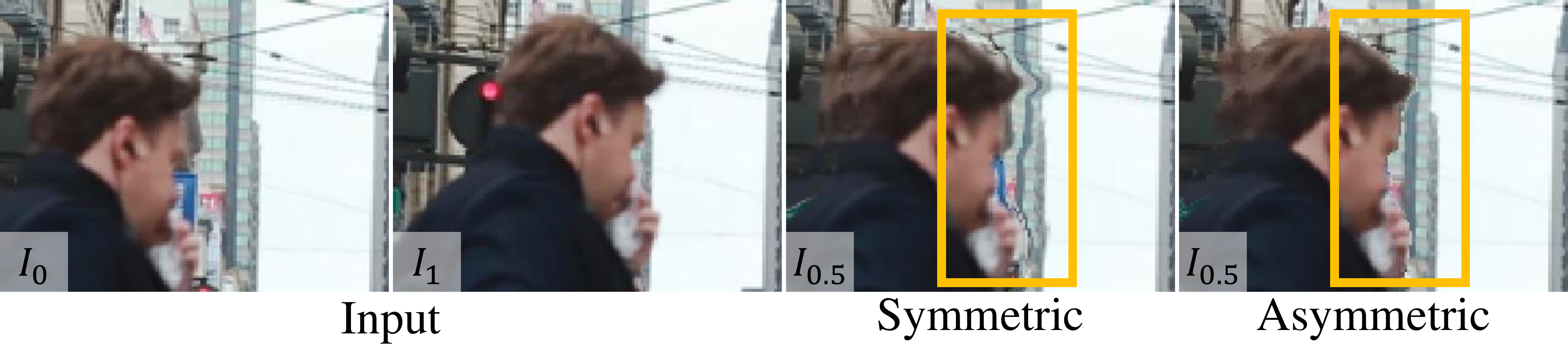}\label{fig:real_example}}
  \caption{Symmetric vs.~asymmetric bilateral motion models. In~(a), the asymmetric model represents bilateral motion vectors from an intermediate frame $I_{0.5}$ to two input frames $I_0$ and $I_1$ accurately, where the symmetric one fails, by loosening the linear constraint. In~(b), when $I_{0.5}$ is interpolated from $I_0$ and $I_1$, the asymmetric model provides more faithful reconstruction with less artifacts, especially around the head, than the symmetric one does.}
  \label{fig:Intro}
  \vspace{-0.2cm}
\end{figure}

On the other hand, many algorithms \cite{choi2007motion, jiang2018slomo, bao2018memc, bao2019dain, niklaus2018ctx, xue2019toflow, liu2019cyclicgen} are based on backward warping, which is free from the hole and overlapping issues. Backward warping needs motion vectors from intermediate frames to input frames, but intermediate frames, which should be interpolated, are unavailable at the time of motion estimation. Thus, conventional algorithms \cite{jiang2018slomo, bao2018memc, bao2019dain} approximate those intermediate motion vectors using optical flows between  input frames. However, the approximation errors may degrade frame interpolation performance. Park~\etal \cite{park2020bmbc} employ the symmetric bilateral motion estimation to improve the accuracy of the intermediate motion, assuming that motion trajectories between input frames are linear. The linear motion constraint, however, may cause inaccurate motion estimation in regions where the constraint is invalid, such as occluded regions around motion boundaries, as illustrated in Figure~\ref{fig:Intro}.

\begin{figure*}[t]
  \centering
  \subfloat [Motion field
  ${\cal V}_{0 \rightarrow 1}$]{\includegraphics[width=0.25\linewidth]{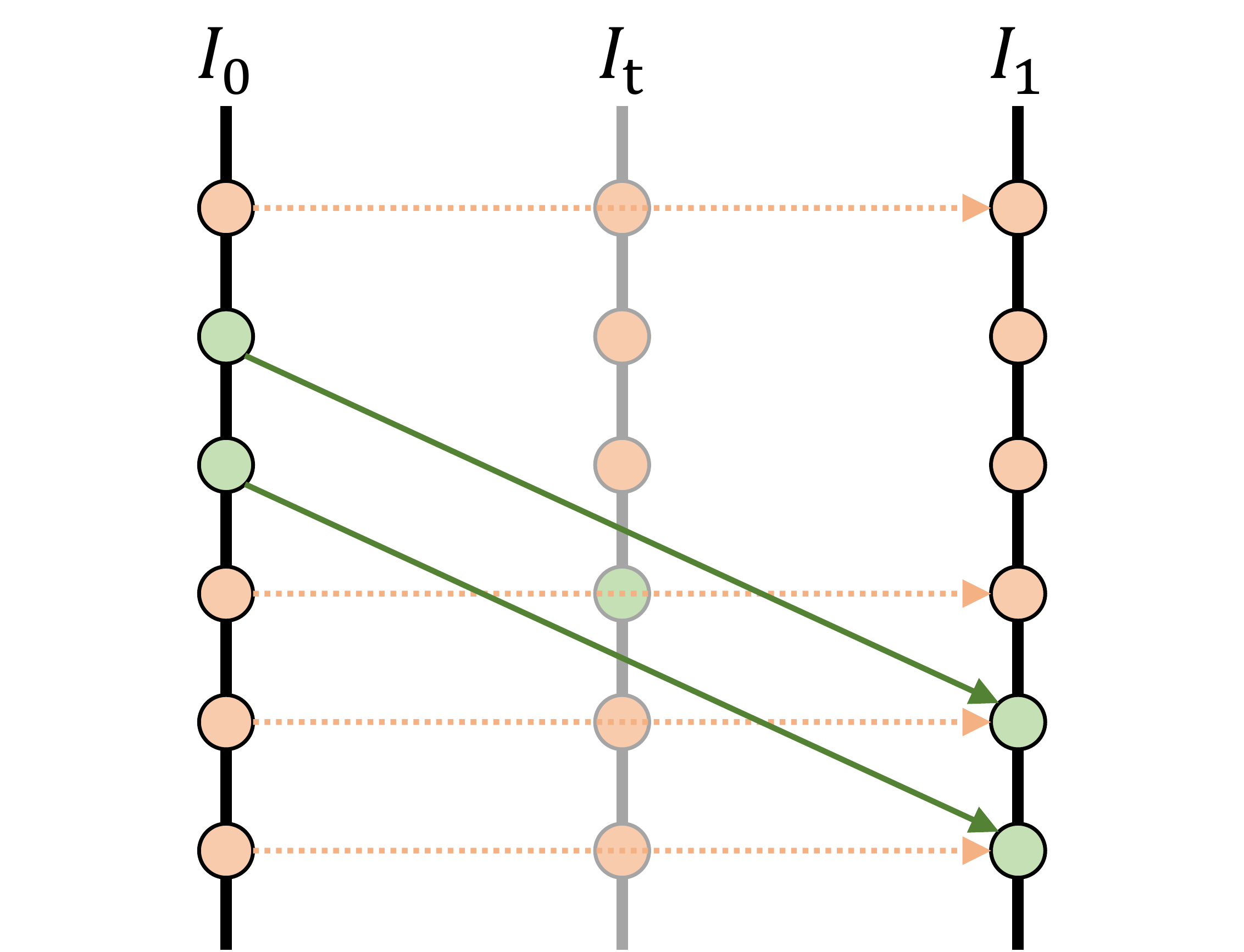}\label{fig:1D_optical_flow}}
  \subfloat [Approximation $\widehat{\cal V}_{t \rightarrow 1}=0.5 \mathcal{V}_{0 \rightarrow 1} $]{\includegraphics[width=0.25\linewidth]{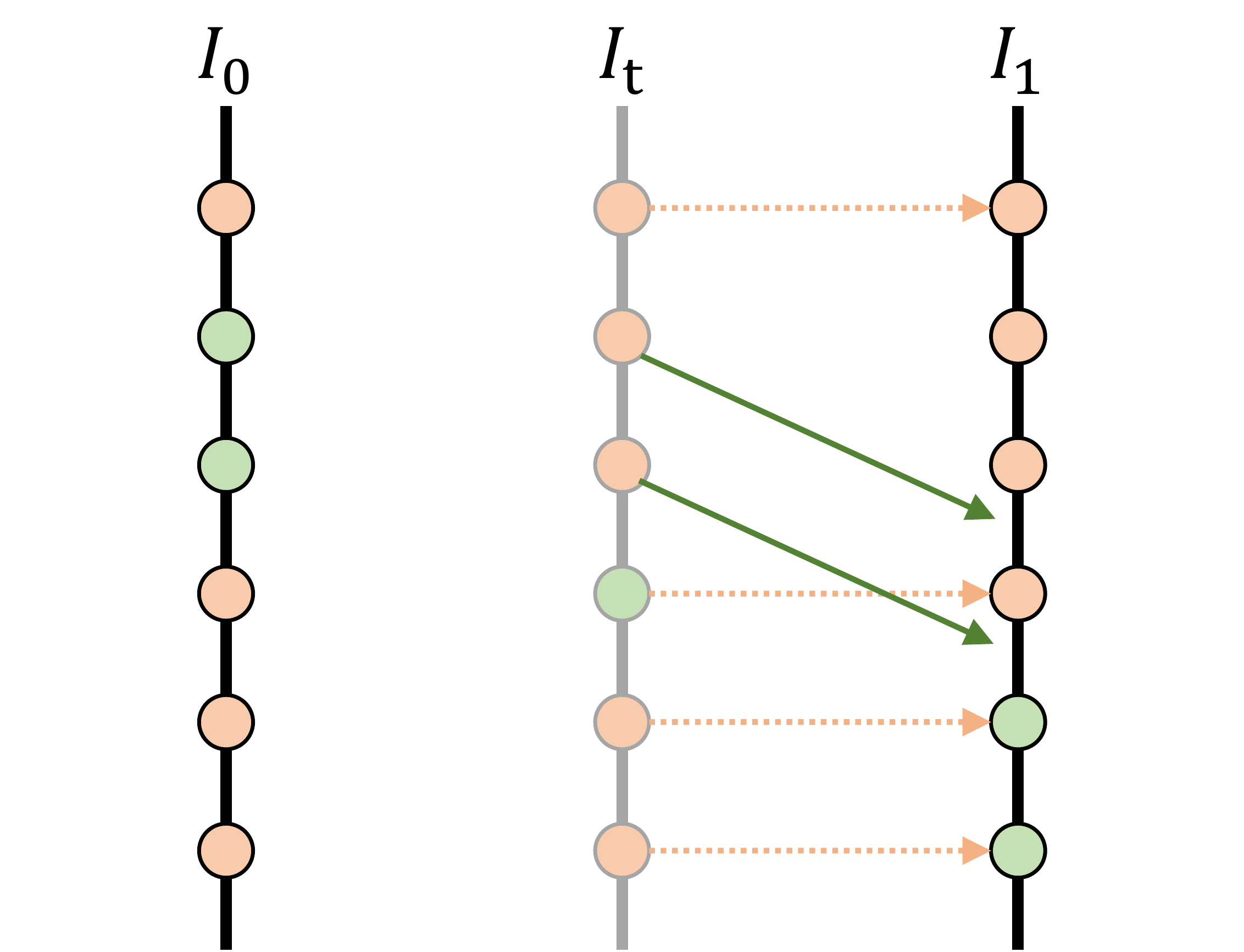}\label{fig:1D_approximation}}
  \subfloat [Symmetric bilateral motion]{\includegraphics[width=0.25\linewidth]{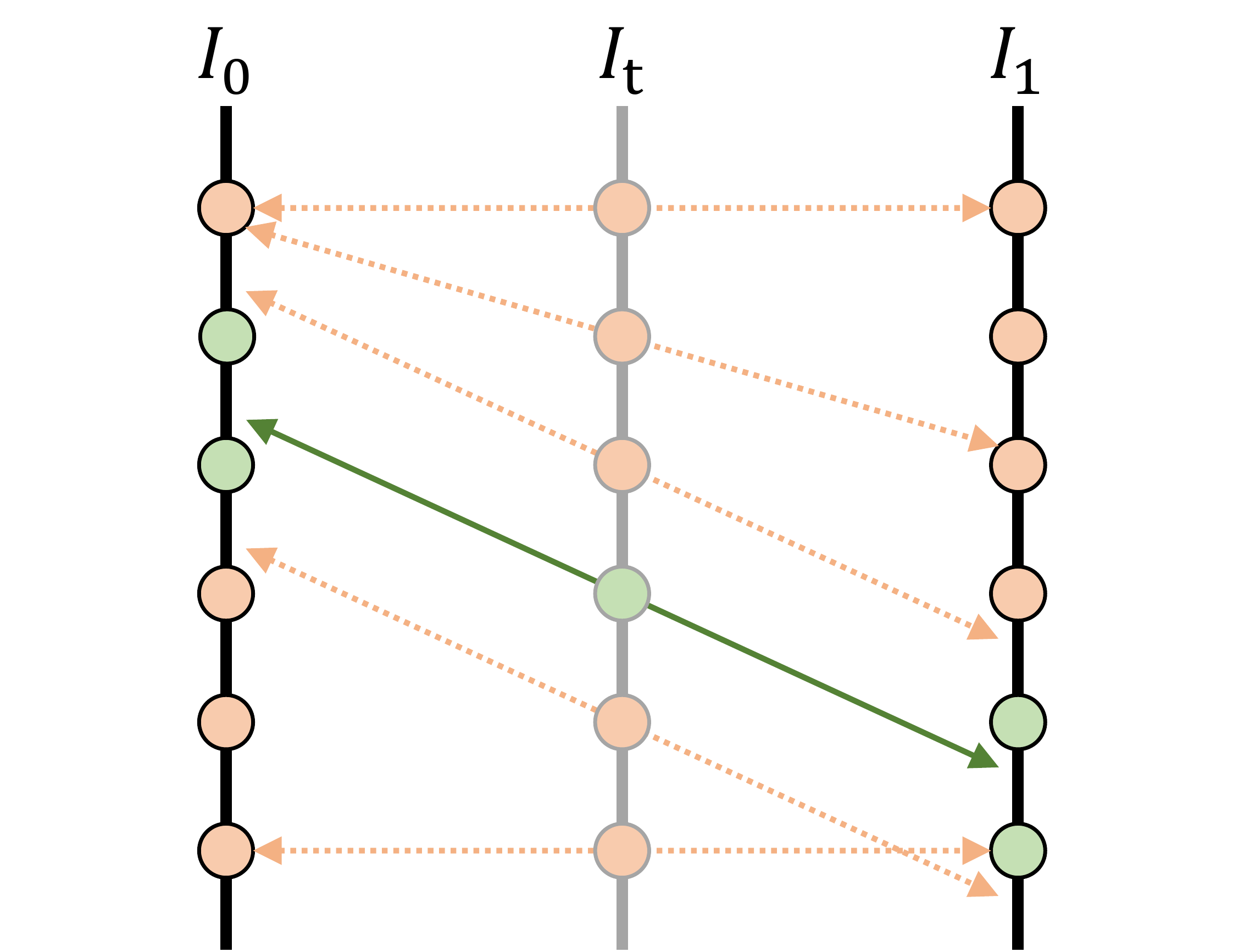}\label{fig:1D_BM}}
  \subfloat [Asymmetric bilateral motion]{\includegraphics[width=0.25\linewidth]{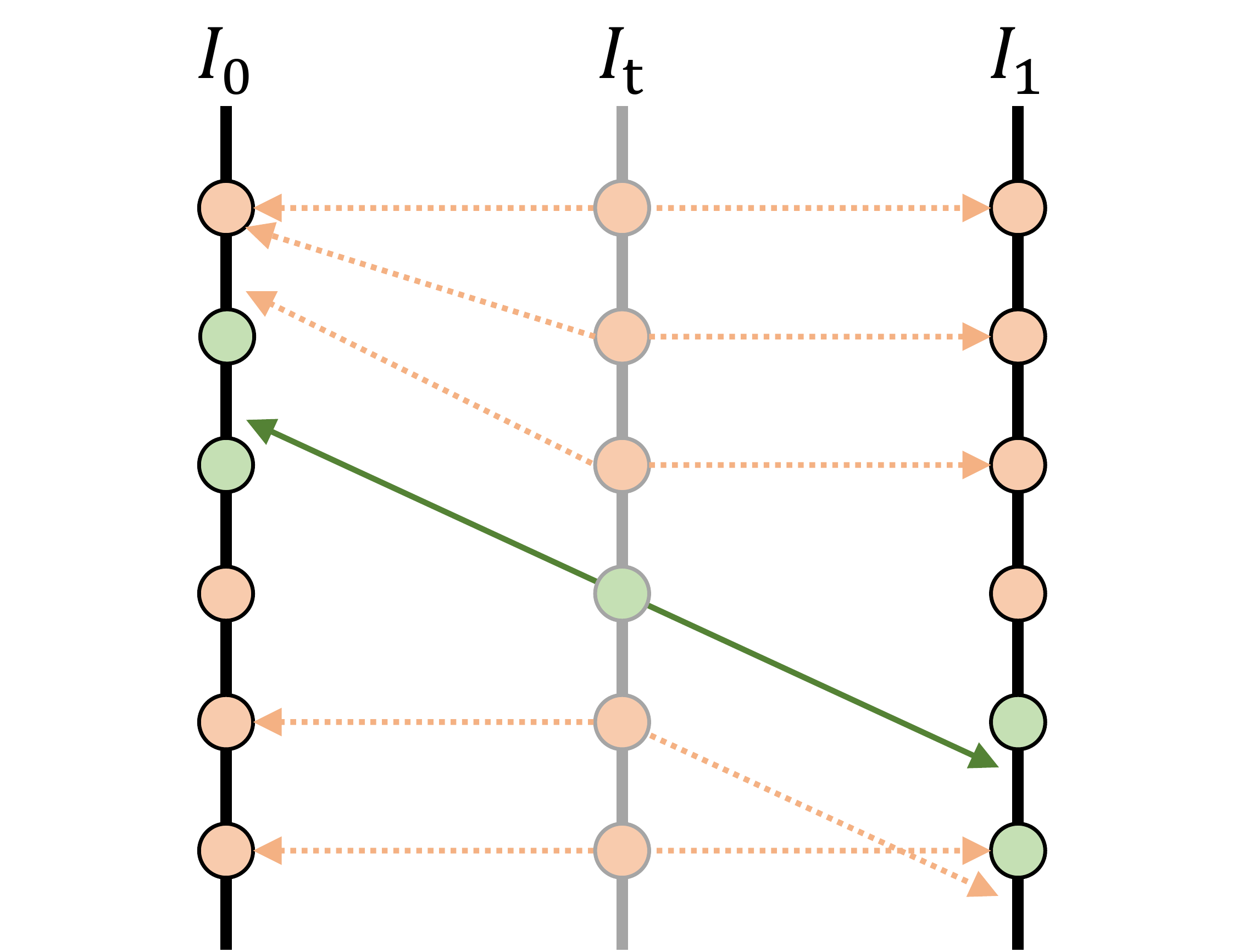}\label{fig:1D_AM}}
  \vspace*{0.1cm}
  \caption{Illustration of various motion fields: Each column represents a frame, and a dot corresponds to a pixel in the frame. $I_0$ and $I_1$ are input frames, and $I_t$ is an unavailable intermediate frame at time instance $t=0.5$. Orange dots depict background pixels without any movement, while green dots depict moving objects. The motion field ${\cal V}_{0 \rightarrow 1}$ in (a) is halved to approximate the motion field $\widehat{\cal V}_{t \rightarrow 1}$ in (b). The symmetric bilateral motion fields in (c) can be estimated to interpolate $I_t$ based on backward warping. To improve the video frame interpolation performance, we propose the ABME algorithm, illustrated in (d). }\label{fig:1D_motions}
  \vspace*{-0.2cm}
\end{figure*}

In this paper, we propose a novel video frame interpolation algorithm based on backward warping, composed of the asymmetric bilateral motion estimation (ABME) and the frame synthesis network. In ABME, we predict symmetric bilateral motion fields and refine them by loosening the linear motion constraint. Specifically, we interpolate a temporary intermediate frame, called an anchor frame, using the symmetric fields. Then, we estimate asymmetric bilateral motion fields from the anchor frame to the two input frames, as illustrated by the red arrows in Figure~\ref{fig:Intro}. In the frame synthesis, the input frames are warped using the bilateral motion fields. To aggregate these warped frames, we develop a synthesis network composed of two subnetworks: FilterNet and RefineNet. FilterNet generates dynamic filters to exploit local information, while RefineNet reconstructs a residual frame using global information. Experimental results demonstrate that the proposed ABME algorithm outperforms the state-of-the-art video interpolators \cite{bao2019dain,niklaus2017sepconv,liu2019cyclicgen,xue2019toflow, choi2020cain, lee2020adacof, park2020bmbc} meaningfully on various  datasets.


\section{Motion-Based Frame Warping}

Let us review motion-based frame warping techniques for video frame interpolation, and introduce the notations and concepts necessary to describe how the proposed ABME is different from the conventional techniques.

\vspace*{0.15cm}
\noindent \textbf{Forward and backward warping:} Let ${\cal V}_{\textrm{S} \rightarrow \textrm{T}}$ denote a pixel-wise motion field (or optical flow) from a source frame $I_\textrm{S}$ to a target frame $I_\textrm{T}$. Then, the target frame can be approximated by forward warping the source frame~\cite{fant1986forwardwarping},
\begin{equation}\label{eq:forward_warping_definition}
\hat{I}_\textrm{T} = \phi_\textrm{F}(I_\textrm{S}, {\cal V}_{\textrm{S} \rightarrow \textrm{T}})
\end{equation}
where $\phi_\textrm{F}$ is the forward warping operator. On the contrary, the source frame can be approximated by backward warping the target frame~\cite{wolberg1990invwarping},
\begin{equation}\label{eq:backward_warping_definition}
\hat{I}_\textrm{S} = \phi_\textrm{B}({\cal V}_{\textrm{S} \rightarrow \textrm{T}}, I_\textrm{T})
\end{equation}
where $\phi_\textrm{B}$ is the backward warping operator.

\begin{figure*}[t]
    \centering
    \includegraphics[width=\linewidth]{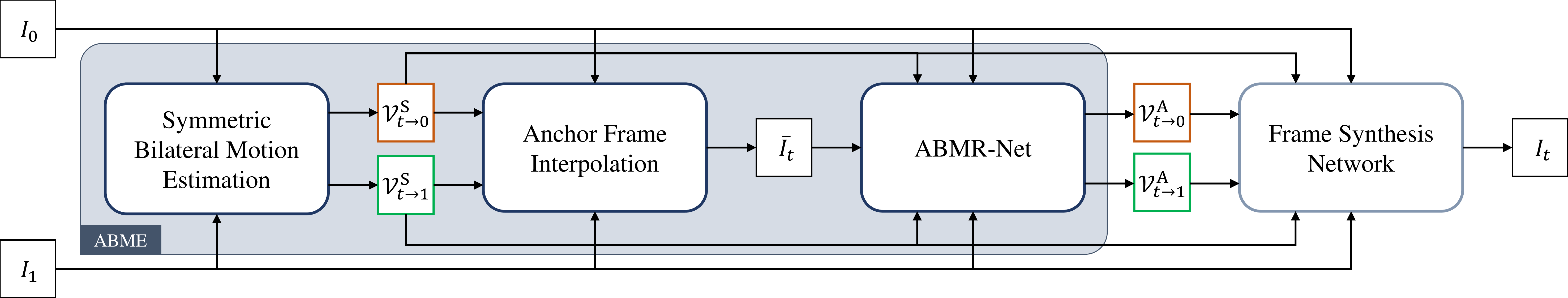}
    \caption{An overview of the proposed algorithm. ABMR-Net is detailed in Figure~\ref{fig:motion_networks}, and the frame synthesis network in Figure~\ref{fig:overview_synthesis}.}
    \label{fig:overview_motion}
    \vspace*{-0.2cm}
\end{figure*}

Given two input frames $I_0$ and $I_1$ at adjacent time instances $0$ and $1$, video frame interpolation aims to synthesize an intermediate frame $I_{t}$, where $0 < t < 1$. This can be achieved, using the forward warping, by
\begin{equation}
\hat{I}_{t, \textrm{F}} = (1-t) \cdot \phi_\textrm{F}(I_\textrm{0}, {\cal V}_{0 \rightarrow t}) + t \cdot \phi_\textrm{F}(I_\textrm{1}, {\cal V}_{1 \rightarrow t}).
\label{eq:f_warp}\
\end{equation}
Here, the required motion fields are often obtained by scaling the motion fields ${\cal V}_{0 \rightarrow 1}$ and ${\cal V}_{1 \rightarrow 0}$ between the input frames~\cite{niklaus2018ctx,niklaus2020softsplatting}, which are given by
\begin{align}
{\cal V}_{0 \rightarrow t} &= t \cdot{\cal V}_{0 \rightarrow 1} \label{eq:f_approx_0} \\
{\cal V}_{1 \rightarrow t} &= (1-t) \cdot {\cal V}_{1 \rightarrow 0} \label{eq:f_approx_1}.
\end{align}
However, as illustrated in Figure~\ref{fig:1D_motions}(a), the scaled ${\cal V}_{0 \rightarrow t}$ does not pass through pixels in $I_t$ exactly in general. Moreover, no flow vector may pass near a certain pixel, or multiple vectors may pass near the same pixel, causing hole or occlusion problems, respectively. Softmax splatting~\cite{niklaus2020softsplatting} alleviates these problems in the forward warping. On the other hand, a majority of video frame interpolation methods \cite{liu2017dvf,jiang2018slomo,bao2018memc,liu2019cyclicgen,bao2019dain,xue2019toflow, Reda2019ucc, park2020bmbc}, as well as the proposed algorithm, instead use the backward warping,
\begin{equation}\label{eq:b_warp}
\hat{I}_{t, \textrm{B}} = (1-t) \cdot \phi_\textrm{B}({\cal V}_{t \rightarrow 0}, I_\textrm{0}) + t \cdot \phi_\textrm{B}({\cal V}_{t \rightarrow 1}, I_\textrm{1}).
\end{equation}
However, unlike \eqref{eq:f_approx_0} and \eqref{eq:f_approx_1}, it is not straightforward to obtain the motion fields ${\cal V}_{t \rightarrow 0}$ and ${\cal V}_{t \rightarrow 1}$ because the intermediate frame $I_t$ is unavailable.

\vspace*{0.15cm}
\noindent \textbf{Motion approximation for backward warping:}
Conventional algorithms~\cite{jiang2018slomo, bao2018memc, bao2019dain,Reda2019ucc,park2020bmbc} approximate the motion fields ${\cal V}_{t \rightarrow 0}$ and ${\cal V}_{t \rightarrow 1}$ in \eqref{eq:b_warp}. For example, the flow projection in \cite{bao2018memc,bao2019dain} approximates ${\cal V}_{t \rightarrow 0}$ and ${\cal V}_{t \rightarrow 1}$ by aggregating multiple flow vectors between $I_0$ and $I_1$, which pass near each pixel in $I_t$. Alternatively, some algorithms simply borrow flow vectors from ${\cal V}_{0 \rightarrow 1}$ and ${\cal V}_{1 \rightarrow 0}$ to approximate ${\cal V}_{t \rightarrow 0}$ and ${\cal V}_{t \rightarrow 1}$ \cite{park2020bmbc},
\begin{align}
\widehat{\cal V}_{t \rightarrow 0} &= - t \cdot {\cal V}_{0 \rightarrow 1} \,\, \textrm{ or } \,\, t \cdot {\cal V}_{1 \rightarrow 0}
\label{eq:m_approx_0} \\
\widehat{\cal V}_{t \rightarrow 1} &= (1-t) \cdot {\cal V}_{0 \rightarrow 1} \,\, \textrm{ or } \,\, -(1-t) \cdot {\cal V}_{1 \rightarrow 0}. \label{eq:m_approx_1}
\end{align}
In \cite{jiang2018slomo, Reda2019ucc}, the motion fields are approximated by combining the candidates in \eqref{eq:m_approx_0} and \eqref{eq:m_approx_1}, given by
\begin{align}
\widehat{\cal V}_{t \rightarrow 0} &= -(1-t)t\cdot{\cal V}_{0 \rightarrow 1} + t^2 \cdot {\cal V}_{1 \rightarrow 0} \label{eq:m_approx_2} \\
\widehat{\cal V}_{t \rightarrow 1} &= (1-t)^2 \cdot{\cal V}_{0 \rightarrow 1} - t(1-t) \cdot {\cal V}_{1 \rightarrow 0}. \label{eq:m_approx_3}
\end{align}

These approximations in \eqref{eq:m_approx_0}$\sim$\eqref{eq:m_approx_3} assume that neighboring pixels have similar motion vectors. However, as in Figure~\ref{fig:1D_motions}(b),  the assumption is invalid around motion boundaries. In such cases, the qualities of the approximate motion fields are degraded, resulting in poorly interpolated frames.

\vspace*{0.15cm}
\noindent \textbf{Symmetric bilateral motion estimation:}
Instead of approximating the bilateral motion fields ${\cal V}_{t \rightarrow 0}$ and ${\cal V}_{t \rightarrow 1}$ using the fields ${\cal V}_{0 \rightarrow 1}$ and ${\cal V}_{1 \rightarrow 0}$ between the input frames, Park \etal~\cite{park2020bmbc} proposed the symmetric bilateral motion estimation algorithm, assuming that a motion trajectory between $I_0$ and $I_1$ is linear. Under the linear assumption, the bilateral motion fields ${\cal V}_{t \rightarrow 0}$ and ${\cal V}_{t \rightarrow 1}$ are symmetric with respect to $I_t$, as in Figure~\ref{fig:1D_motions}(c). More specifically,
\begin{equation}
{\cal V}_{t \rightarrow 0} = -\frac{t}{1-t} {\cal V}_{t \rightarrow 1}
\label{eq:symmetry_condition}.
\end{equation}
Therefore, roughly speaking, they obtained ${\cal V}_{t \rightarrow 1}$ to minimize the frame difference
\begin{eqnarray}
\lefteqn{\| \phi_\textrm{B}({\cal V}_{t \rightarrow 0}, I_\textrm{0}) - \phi_\textrm{B}({\cal V}_{t \rightarrow 1}, I_\textrm{1}) \| = } \nonumber \\
& & \quad \| \phi_\textrm{B}(\textstyle -\frac{t}{1-t} {\cal V}_{t \rightarrow 1}, I_\textrm{0}) - \phi_\textrm{B}({\cal V}_{t \rightarrow 1}, I_\textrm{1}) \|.
\label{eq:SBME}
\end{eqnarray}
To this end, they developed the bilateral motion network with the bilateral cost volume.

\section{Proposed Algorithm}

The proposed algorithm is composed of two procedures: ABME and frame synthesis.

\subsection{ABME}

In Figure~\ref{fig:1D_motions}(c), bilateral motion vectors convey valid motion information between symmetrically matched pixel pairs in input frames. However, when a pixel in $I_t$ is occluded in either $I_0$ or $I_1$, the symmetry does not hold. Nonlinear object motions due to acceleration also break the symmetry, as in Figure~\ref{fig:Intro}(a). To overcome these issues, we develop the ABME technique, which refines symmetric bilateral motion vectors so that they become asymmetric and represent motion more reliably and more accurately.

Figure~\ref{fig:overview_motion} presents an overview of the proposed algorithm. We first obtain symmetric bilateral motion fields $\mathcal{V}^{\textrm{S}}_{t\rightarrow 0}$ and $\mathcal{V}^{\textrm{S}}_{t\rightarrow 1}$ by employing the motion estimator of BMBC~\cite{park2020bmbc}.
Using $\mathcal{V}^{\textrm{S}}_{t\rightarrow 0}$ and $\mathcal{V}^{\textrm{S}}_{t\rightarrow 1}$, we interpolate an anchor frame $\bar{I}_t$, which is then used as a source frame for the asymmetric bilateral motion refinement (ABMR). Finally, we obtain asymmetric bilateral motion fields $\mathcal{V}^{\textrm{A}}_{t\rightarrow 0}$ and $\mathcal{V}^{\textrm{A}}_{t\rightarrow 1}$.

\vspace*{0.15cm}
\noindent \textbf{Anchor frame interpolation:} The motion estimation from a source frame $I_t$ to a target frame $I_0$ or $I_1$ is challenging, since $I_t$ is unavailable and should be synthesized in the video frame interpolation. Hence, we generate a temporary source frame $\bar{I}_t$, called an anchor frame, using the symmetric bilateral motion fields $\mathcal{V}^{\textrm{S}}_{t\rightarrow 0}$ and $\mathcal{V}^{\textrm{S}}_{t\rightarrow 1}$.

Based on the backward warping in \eqref{eq:b_warp}, we may generate
\begin{equation}\label{Eq:anchor_frame_v1}
  \bar{I}_t = (1-t)\cdot\phi_\textrm{B}({\cal V}^{\textrm{S}}_{t \rightarrow 0}, I_0) + t\cdot\phi_\textrm{B}({\cal V}^{\textrm{S}}_{t \rightarrow 1}, I_1).
\end{equation}
This simple blending, however, may cause errors due to occlusion, especially in boundary regions of the anchor frame in the case of camera panning. To reduce such errors, we exploit masks to reveal occluded regions, given by
\begin{equation}\label{Eq:warping_mask}
  {M}^{\textrm{S}}_{t \rightarrow 0} = \phi_\textrm{B}({\cal V}^{\textrm{S}}_{t \rightarrow 0}, \mathbf{1})
  \,\, \textrm{ and } \,\,
  {M}^{\textrm{S}}_{t \rightarrow 1} = \phi_\textrm{B}({\cal V}^{\textrm{S}}_{t \rightarrow 1}, \mathbf{1})
\end{equation}
where $\mathbf{1}$ is a binary image of all ones. Note that a mask value 0 means that the corresponding pixel in $I_t$ moves outside the frame at time instance $0$ or $1$. We then reconstruct the anchor frame in an occlusion-aware manner,
\begin{align}
        \bar{I}_t &= (1-t)\cdot(\mathbf{1}-{M}^{\textrm{S}}_{t \rightarrow 1}+{M}^{\textrm{S}}_{t \rightarrow 0}) \otimes \phi_\textrm{B}({\cal V}^{\textrm{S}}_{t \rightarrow 0}, I_0) \nonumber \\
        & \quad + t\cdot(\mathbf{1}-{M}^{\textrm{S}}_{t \rightarrow 0}+{M}^{\textrm{S}}_{t \rightarrow 1}) \otimes \phi_\textrm{B}({\cal V}^{\textrm{S}}_{t \rightarrow 1}, I_1)   \label{Eq:anchor_frame_v2}
\end{align}
where $\otimes$ is the Hadamard product.

\begin{figure*}[t]
    \centering
    \includegraphics[width=\linewidth]{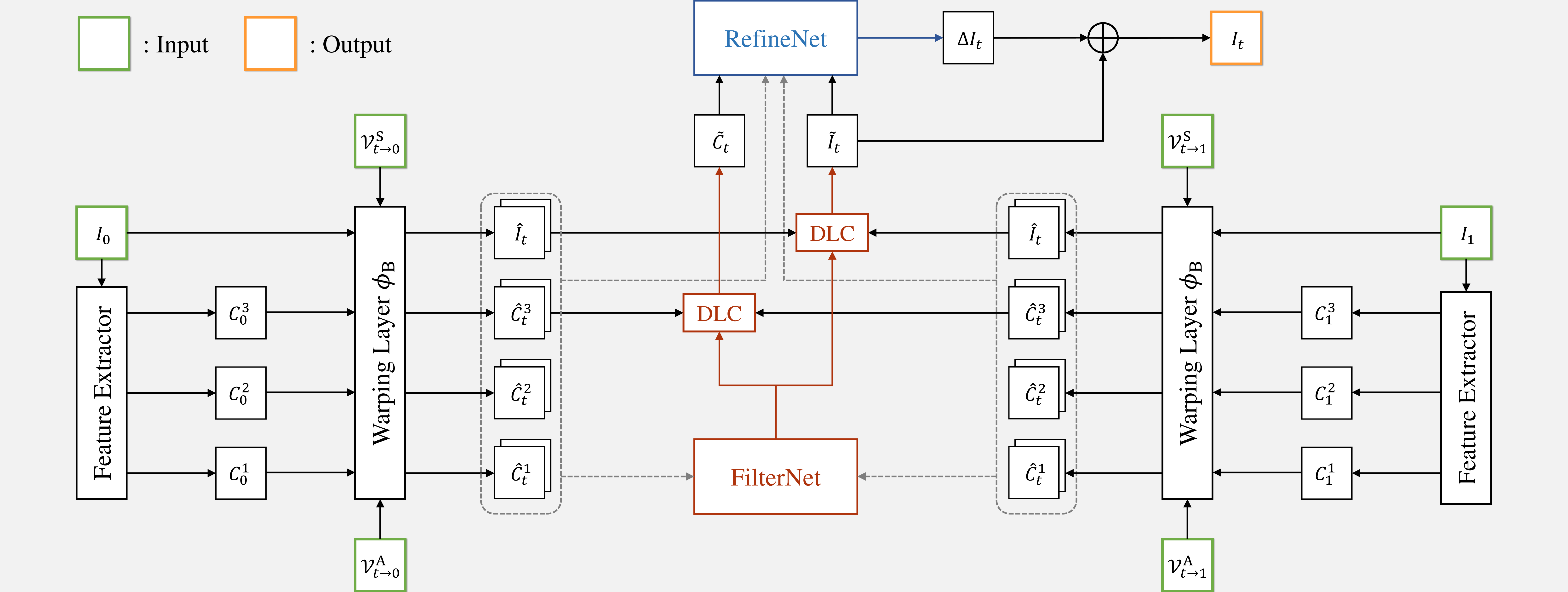}
    \caption{Architecture of the proposed frame synthesis network.}
    \label{fig:overview_synthesis}
    \vspace*{-0.2cm}
\end{figure*}

\begin{figure}
  \centering
  \includegraphics[width=\linewidth]{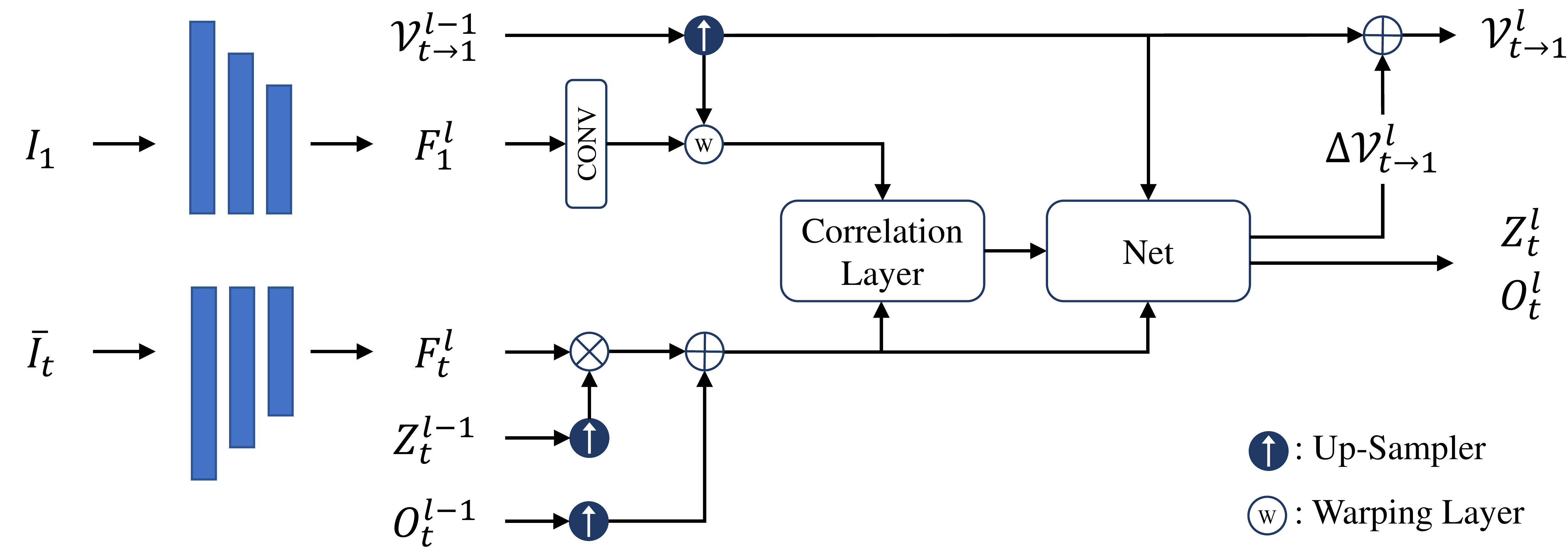}
  \vspace*{0.01cm}
  \caption{Architecture of ABMR-Net.}
  \label{fig:motion_networks}
  \vspace*{-0.3cm}
\end{figure}

\vspace*{0.15cm}
\noindent \textbf{Asymmetric bilateral motion refinement:} To perform the backward warping in \eqref{eq:b_warp}, conventional algorithms approximate bilateral motion fields ${\cal V}_{t \rightarrow 0}$ and ${\cal V}_{t \rightarrow 1}$ using the motion fields between $I_0$ and $I_1$. In contrast, we estimate the motion field from $I_t$ to $I_0$ or $I_1$ directly, after approximating $I_t$ with the anchor frame $\bar{I}_t$.

Let us describe the asymmetric motion estimation from $\bar{I}_t$ to $I_1$. Note that the estimation from  $\bar{I}_t$ to $I_0$ is performed similarly but independently. We develop ABMR-Net for asymmetric bilateral motion refinement in Figure~\ref{fig:motion_networks} to refine the motion field from the source $\bar{I}_t$ to the target $I_1$. ABMR-Net hierarchically obtains the motion field, as done in PWC-Net~\cite{sun2018pwc}. At level $l$, the motion field $\mathcal{V}^{l-1}_{t \rightarrow 1}$ at the previous level $(l-1)$ is up-sampled to warp the target feature map $F^l_1$. Also, we multiply the anchor feature map $F^l_t$ with a reliability mask $Z^{l-1}_t$ and compensate for masked-out features by adding an offset map $O^{l-1}_t$. Then, the warped target feature map and the compensated anchor feature map are input to the correlation layer to compute matching costs. The cost volume is used to generate the residual field $\Delta \mathcal{V}^{l}_{t \rightarrow 1}$, which is added to the up-sampled $\mathcal{V}^{l-1}_{t \rightarrow 1}$ to yield the motion field~$\mathcal{V}^{l}_{t \rightarrow 1}$.

As mentioned previously, the symmetric field ${\cal V}^{\textrm{S}}_{t \rightarrow 1}$ is estimated using the motion estimator of BMBC~\cite{park2020bmbc}, which is at the quarter resolution. It is used as the up-sampled $\mathcal{V}^{0}_{t \rightarrow 1}$ at level $l=1$ in Figure~\ref{fig:motion_networks}. Then, the refinement is performed for two levels, and the half resolution $\mathcal{V}^{2}_{t \rightarrow 1}$ becomes the final result ${\cal V}^{\textrm{A}}_{t \rightarrow 1}$. Since ${\cal V}^{\textrm{A}}_{t \rightarrow 0}$ and ${\cal V}^{\textrm{A}}_{t \rightarrow 1}$ are refined separately, they become asymmetric. As shown in Figure~\ref{fig:Intro}(a) and Figure~\ref{fig:1D_motions}(d), the asymmetric fields may represent motion information more faithfully than the symmetric ones.

Recently, Zhao \etal~\cite{zhao2020maskflownet} improved the matching performance of PWC-Net with a learnable occlusion mask. Similarly, we employ a mask in ABMR-Net to improve the refinement performance. The source frame $\bar{I}_t$ is an approximation of $I_t$, so it may contain errors around motion boundaries or on complicated texture. Such errors make the matching process unreliable. Hence, we adopt the reliability mask $Z^{l-1}_t$ in Figure~\ref{fig:motion_networks} to suppress errors in anchor features. Notice that a source frame is masked in this work, while a target frame is masked in~\cite{zhao2020maskflownet}. At level 1, $O_t^{0}$ is initialized to zero, and the up-sampled $Z_t^{0}$ is set to
\begin{equation}\label{Eq:Initial_reliability_map}
  \uparrow Z^{0}_t  = \exp\!\Big(-\beta\cdot \big|\phi_\textrm{B}({\cal V}^{\textrm{S}}_{t \rightarrow 0}, I_0) - \phi_\textrm{B}({\cal V}^{\textrm{S}}_{t \rightarrow 1},I_1)\big| \Big)
\end{equation}
where $\beta=20$. If a certain pixel in $\bar{I}_t$ has a large symmetric matching error in $|\phi_\textrm{B}({\cal V}^{\textrm{S}}_{t \rightarrow 0}, I_0) - \phi_\textrm{B}({\cal V}^{\textrm{S}}_{t \rightarrow 1},I_1)|$, its feature is suppressed by the reliability mask.

\subsection{Frame Synthesis}

In Figure~\ref{fig:overview_synthesis}, we synthesize an intermediate frame $I_t$ using two input frames $I_0$ and $I_1$. We use four motion fields generated by the proposed ABME in Figure~\ref{fig:overview_motion}: two symmetric fields ${\cal V}^{\textrm{S}}_{t \rightarrow 0}$ and ${\cal V}^{\textrm{S}}_{t \rightarrow 1}$ and two asymmetric fields ${\cal V}^{\textrm{A}}_{t \rightarrow 0}$ and ${\cal V}^{\textrm{A}}_{t \rightarrow 1}$.


Using a feature extractor for frame synthesis, we extract multi-scale feature maps $C_0^l$ from $I_0$, where there are three levels $l\in \{1, 2, 3\}$. Note that the extraction from  $I_1$ to $C_1^l$ is performed similarly with shared parameters. The highest level maps $C_0^3$ and $C_1^3$ have the same spatial resolution as the input frames. We backward warp the input images and their feature pyramids. In Figure~\ref{fig:overview_synthesis}, $I_0$ is warped by ${\cal V}^{\textrm{S}}_{t \rightarrow 0}$ and ${\cal V}^{\textrm{A}}_{t \rightarrow 0}$ to obtain the estimate $\hat{I}_t$ of $I_t$, while $I_1$ by ${\cal V}^{\textrm{S}}_{t \rightarrow 1}$ and ${\cal V}^{\textrm{A}}_{t \rightarrow 1}$. Thus, there are four candidate warped frames $\hat{I}_t$ in total. Similarly, at each level $l$, there are four candidate warped features $\widehat{C}_t^l$. These candidates are combined in a complementary manner to reconstruct the intermediate frame $I_t$ more faithfully.

Given multiple candidates, we can synthesize $I_t$ by blending them with weights. This simple blending, however, may cause blurry artifacts and reconstruction errors in occluded regions. Recent algorithms hence employ synthesis networks, which process warped frames to generate the intermediate frame in various ways: direct frame generation~\cite{niklaus2018ctx, niklaus2020softsplatting}, residual frame generation~\cite{bao2019dain, bao2018memc}, or dynamic local blending~\cite{park2020bmbc}. While the dynamic local blending synthesizes each pixel using local neighbors, the other two approaches use global contexts. To exploit both local and global information, we propose a novel synthesis network composed of two subnetworks: FilterNet and RefineNet.

\begin{figure}
  \centering
  \includegraphics[width=0.98\linewidth]{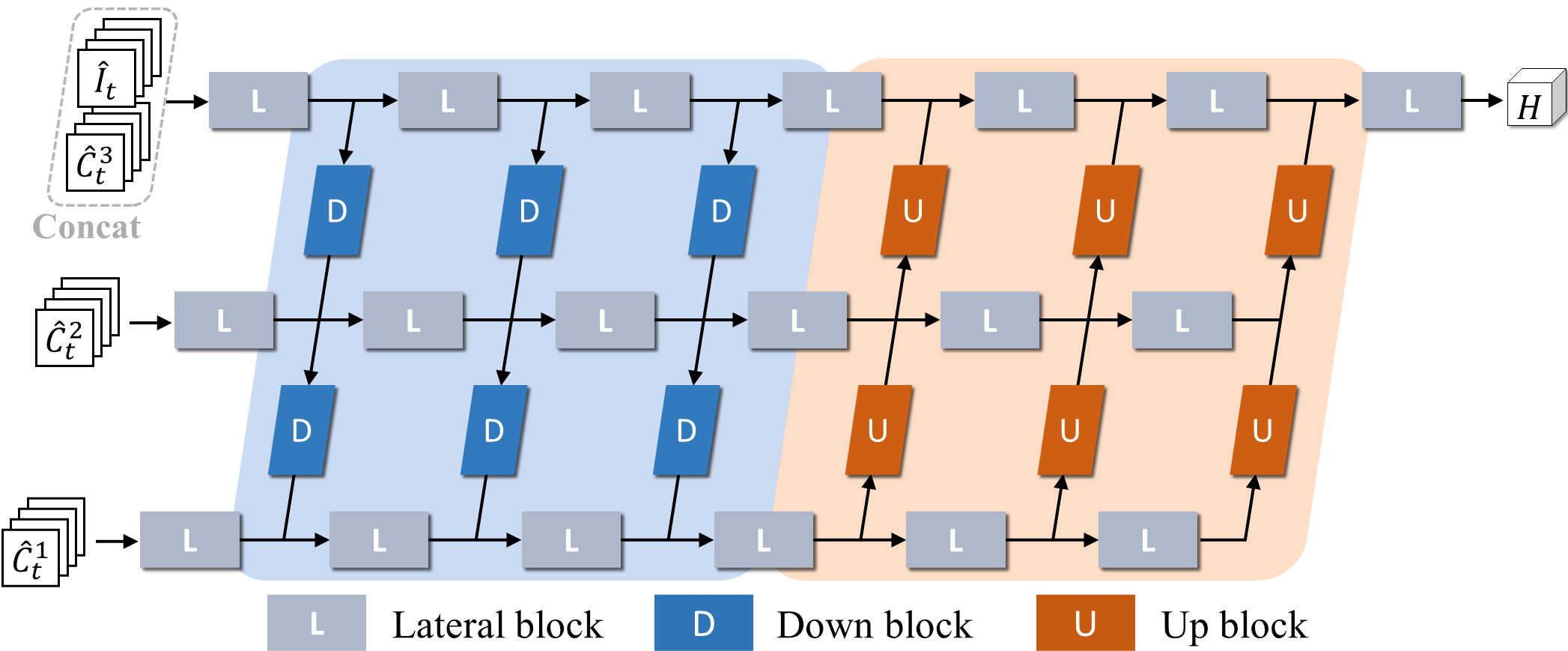}
  \vspace*{0.01cm}
  \caption{Modified GridNet for FilterNet}\label{fig:grid Net}
  \vspace*{-0.2cm}
\end{figure}

\vspace*{0.15cm}
\noindent \textbf{FilterNet:} It learns to generate dynamic filters for combining the four candidates, denoted by $\hat{I}_t^c$, $1\leq c \leq 4$.
We employ the modified version~\cite{niklaus2018ctx} of GridNet~\cite{fourure2017gridnet} as the backbone of FilterNet, which is shown in Figure~\ref{fig:grid Net}. FilterNet takes the four candidates with the corresponding feature pyramids as input, by employing the lateral blocks.
For each pixel $(x,y)$, the rightmost lateral block generates filter coefficients dynamically to fuse the $3\times 3$ local neighboring pixels in each candidate. The coefficients are denoted by
\begin{equation}
H_{x,y}(i,j,c), \quad -1 \leq i,j \leq 1, \quad 1 \leq c \leq 4
\end{equation}
where $(i,j)$ are local coordinates around $(x,y)$, and $c$ is the candidate index. The filter coefficients are normalized, $\sum_c \sum_i \sum_j H_{x,y}(i,j,c) = 1$. Then, we obtain the filtered frame via the dynamic local convolution (DLC),
\begin{equation}\label{eq:Local filtering}
  \tilde{I}_t(x,y)=\sum_{c=1}^{4} \sum_{i=-1}^{1}  \sum_{j=-1}^{1} H_{x,y}(i,j,c) \hat{I}^c_t(x+i,y+j).
\end{equation}
Furthermore, by applying the same dynamic filters to the four warped feature candidates $\widehat{C}^l_t$ at the highest level $l=3$, we obtain the filtered feature map $\widetilde{C}_t$.

\begin{table*}[t]
    \renewcommand{\arraystretch}{0.95}
    \caption
   {
        Quantitative comparison (PSNR/SSIM) of video interpolation results on the UCF101, Vimeo90K, and SNU-FILM datasets. In each test, the best result is \textcolor{red}{\textbf{boldfaced}}, while the second best is \textcolor{blue}{\underline{underlined}}. All results are obtained by executing available source codes.
    }
    \centering
    {\footnotesize
    \begin{tabular}{L{2.20cm}C{1.5cm}C{1.5cm}C{1.5cm}C{1.5cm}C{1.5cm}C{1.5cm}C{1cm}C{1.3cm}}
    \toprule
    \multirow{2}[2]{*}{} & \multirow{2}[1]{*}{UCF101} & \multirow{2}[1]{*}{Vimeo90K} & \multicolumn{4}{c}{SNU-FILM} & \multirow{2}[2]{*}{\makecell{Runtime \\ (seconds)}} & \multirow{2}[2]{*}{\makecell{\#Parameters \\ (millions)}} \\[-0.1em]
    \cmidrule(lr){4-7}
    & & & Easy& Medium&Hard&Extreme& &\\
    \midrule
    ToFlow\cite{xue2019toflow}               &34.58/0.9667 &33.73/0.9682 &39.08/0.9890 &34.39/0.9740 &28.44/0.9180 &23.39/0.8310 &0.43 & 1.1\\[0.1em]
    SepConv\cite{niklaus2017sepconv}         &34.78/0.9669 &33.79/0.9702 &39.41/0.9900 &34.97/0.9762 &29.36/0.9253 &24.31/0.8448&0.20 & 21.6\\[0.1em]
    CyclicGen\cite{liu2019cyclicgen}         &35.11/0.9684 &32.09/0.9490 &37.72/0.9840 &32.47/0.9554 &26.95/0.8871 &22.70/0.8083&0.09 & 19.8\\[0.1em]
    DAIN\cite{bao2019dain}                   &34.99/0.9683 &34.71/0.9756 &39.73/\textcolor{red}{\bf 0.9902} &35.46/\textcolor{blue}{\underline{0.9780}} &\textcolor{blue}{\underline{30.17}}/\textcolor{blue}{\underline{0.9335}} &\textcolor{blue}{\underline{25.09}}/\textcolor{blue}{\underline{0.8584}} &0.13 & 24.0\\[0.1em]
    CAIN\cite{choi2020cain}                  &34.91/\textcolor{blue}{\underline{0.9690}} &34.65/0.9730 &\textcolor{blue}{\underline{39.89}}/0.9900 &\textcolor{blue}{\underline{35.61}}/0.9776 &29.90/0.9292 &24.78/0.8507 &0.04 & 42.8\\[0.1em]
    AdaCoF\cite{lee2020adacof}               & 34.90/0.9680 &34.47/0.9730 &39.80/0.9900 &35.05/0.9754 &29.46/0.9244 &24.31/0.8439 &0.03 & 22.9\\[0.1em]
    BMBC\cite{park2020bmbc}                  &\textcolor{blue}{\underline{35.15}}/0.9689 &\textcolor{blue}{\underline{35.01}}/\textcolor{blue}{\underline{0.9764}} &\textcolor{red}{\bf 39.90}/\textcolor{red}{\bf 0.9902} &35.31/0.9774 &29.33/0.9270 &23.92/0.8432 &0.77 & 11.0\\[0.1em]
    ABME (Proposed)                          &\textcolor{red}{\bf 35.38}/\textcolor{red}{\bf 0.9698} &\textcolor{red}{\bf 36.18}/\textcolor{red}{\bf 0.9805} & 39.59/\textcolor{blue}{\underline{0.9901}} &\textcolor{red}{\bf 35.77}/\textcolor{red}{\bf 0.9789} &\textcolor{red}{\bf 30.58}/\textcolor{red}{\bf 0.9364} &\textcolor{red}{\bf 25.42}/\textcolor{red}{\bf 0.8639} &0.22 & 18.1\\
    \bottomrule\\[-2em]
    \end{tabular}
    }
    \label{table:Evaluation on test set}
\end{table*}

\vspace*{0.15cm}
\noindent \textbf{RefineNet:} The dynamic filters consider local neighbors only. Thus, if the local neighbors do not contain proper information on a certain pixel due to motion errors or severe occlusion, its filtered result becomes also erroneous. To overcome this limitation using global information, RefineNet generates a residual frame $\Delta I_t$ to refine the filtered frame $\tilde{I}_t$. It has the same network architecture as FilterNet, but it takes the filtered frame $\tilde{I}_t$, the filtered feature map $\widetilde{C}_t$, and the warped feature pyramids as input. In particular, the feature map $\widetilde{C}_t$, which is obtained using the same dynamic filters for $\tilde{I}_t$, meaningfully increases the refinement performance, as will be discussed in Section~\ref{sec:exp}. After generating the residual $\Delta I_t$, the final reconstructed frame is given by
\begin{equation}\label{eq:Residual_refinement}
  I_t=\tilde{I}_t+\Delta I_t.
\end{equation}


\section{Experiments}
\label{sec:exp}

\subsection{Training}

We first train the symmetric bilateral motion estimator and then, after fixing it, train ABMR-Net. Finally, we end-to-end train the frame synthesis network with those two networks.

\vspace*{0.15cm}
{\noindent \bf ABME:} We adopt the motion estimator of BMBC~\cite{park2020bmbc} for the symmetric bilateral motion estimation in Figure~\ref{fig:overview_motion}, but we retrain it by setting the stride of the first convolution layer to 2 for efficiency.
To train the motion estimators of both BMBC and ABMR-Net, we define the photometric loss between a ground-truth $I^{\textrm{GT}}_{t}$ and two warped frames as
\begin{align}
    &\mathcal{L}_\textrm{pho} =  \rho(I^{\textrm{GT}}_{t}-\phi_\textrm{B}(\mathcal{V}^{\textrm{A}}_{t\rightarrow 0}, I_0)) + \rho(I^{\textrm{GT}}_{t}-\phi_\textrm{B}(\mathcal{V}^{\textrm{A}}_{t\rightarrow 1}, I_1) \nonumber\\
    &\quad\;\; + \mathcal{L}_\textrm{cen}(I^\textrm{GT}_t,\phi_\textrm{B}(\mathcal{V}^{\textrm{A}}_{t\rightarrow 0}, I_0)) + \mathcal{L}_\textrm{cen}(I^\textrm{GT}_t,\phi_\textrm{B}(\mathcal{V}^{\textrm{A}}_{t\rightarrow 1}, I_1))\nonumber
    \label{Eq:photometric_loss}
\end{align}
where $\rho(x)=(x^2+\epsilon^2)^{\alpha}$ is the Charbonnier function~\cite{Char1994loss} and $\mathcal{L}_\textrm{cen}$ is the census loss~\cite{Meister2018unflow, zhong2019unsupervised, zou2018DF} defined as the soft Hamming distance between census-transformed image patches~\cite{Meister2018unflow} of size $7\times 7$. The parameters are set to $\alpha=0.5$ and $\epsilon=10^{-6}$.

To train the motion estimator of BMBC, we use the Adam optimizer~\cite{kingma2014adam} with a learning rate of $\eta = 10^{-4}$ until 0.1M iterations and halve $\eta$ after every 0.04M iterations. We use a batch size of 24 for 0.2M iterations.
For ABMR-Net, we also use the Adam optimizer with $\eta = 10^{-4}$ until 0.12M iterations and halve $\eta$ after every 0.06M iterations. We use a batch size of 16 for 0.3M iterations.


\vspace*{0.15cm}
{\noindent \bf Frame synthesis network:}
We define the synthesis loss $\mathcal{L}_\textrm{syn}$ as the sum of the Charbonnier loss and the census loss between $I^\textrm{GT}_t$ and its synthesized version $I_t$, given by
\begin{equation}
    \label{eq:reconstruction loss}
    \mathcal{L}_\textrm{syn} = \rho(I^\textrm{GT}_t-I_t) + \mathcal{L}_\textrm{cen}(I^\textrm{GT}_t,I_t).
\end{equation}
We use the Adam optimizer with $\eta = 10^{-4}$ until 0.35M iterations and halve $\eta$ after every 0.15M iterations. We use a batch size of 6 for 0.8M iterations in total.

\vspace*{0.15cm}
{\noindent \bf Training dataset:}
We use only the Vimeo90K training set~\cite{xue2019toflow} to train the proposed networks. It is composed of 51,312 triplets with a resolution of 448$\times$256. This training set does not contain motion ground-truth. We augment the dataset by randomly flipping, rotating, reversing the sequence order, and cropping 256$\times$256 patches.

\subsection{Datasets}
While we use strictly a single training dataset, we test the proposed ABME algorithm on various datasets.

\vspace*{0.15cm}
{\noindent \textbf{UCF101 \cite{soomro2012ucf}}:}
We use the test set constructed by Liu~\etal~\cite{liu2017dvf}, which contains 379 triplets of resolution $256\times256$.

\vspace*{0.15cm}
{\noindent \textbf{Vimeo90K \cite{xue2019toflow}}:}
The test set in Vimeo90K contains 3,782 triplets of spatial resolution $448 \times 256$.

\vspace*{0.15cm}
{\noindent \textbf{SNU-FILM \cite{choi2020cain}}:}
It contains 1,240 triplets of videos of resolutions up to $1280\times720$. It has four different settings -- Easy, Medium, Hard, and Extreme.

\vspace*{0.15cm}
{\noindent \textbf{Xiph \cite{Montgomery1994Xiph}}:}
It contains 30 raw video sequences for testing video codecs that have HD ($1280\times 720$) or FHD ($1920\times 1080$) resolutions. For triplets from the FHD sequences, we crop the central HD part of each frame without resizing. Thus, we extract 2,000 triplets of the HD resolution in total.

We divide these triplets into 5 classes (D1 $\sim$ D5) according to the difficulty levels of interpolation. To quantify the difficulty, we compress each triplet using the recent video coding standard VVC~\cite{Bross2021VVC} with a fixed QP. The middle frame is encoded in the B-frame mode, while the other two frames in the I-frame mode. The B-frame is motion-compensated from the others, and the motion vectors and compensation errors are encoded. The number of bits for the B-frame, hence, represents how difficult it is to interpolate it using the adjacent frames. The 2,000 triplets are sorted according to these numbers of bits and then grouped into the five classes so that each class contains 400 triplets. D1 is the easiest class, while D5 is the most difficult one.

\vspace*{0.15cm}
{\noindent \textbf{X4K1000FPS~\cite{XVFI}}:}
Concurrently with this paper in ICCV 2021, Sim \etal present a high-quality, extensive dataset of 4K resolution. They provide the training set X-TRAIN and the test set X-TEST. We evaluate the proposed algorithm directly on X-TEST without retraining it on X-TRAIN.


\begin{table}[t]
    \vspace*{0.1cm}
    \renewcommand{\arraystretch}{0.95}
    \caption
   {
        PSNRs on the Xiph dataset according to the difficulty levels. D1 is the easiest class, while D5 is the most difficult one.
   }
    \centering
    {\footnotesize
    \begin{tabular}{L{2.3cm}C{0.65cm}C{0.65cm}C{0.65cm}C{0.65cm}C{0.65cm}}
    \toprule
    & D1 & D2 & D3 & D4 & D5\\
    \midrule
    DAIN\cite{bao2019dain}	         &34.65	&\textcolor{blue}{\underline{33.21}}	&\textcolor{blue}{\underline{29.42}}	&\textcolor{blue}{\underline{25.41}}	&22.61 \\[0.1em]
    CAIN\cite{choi2020cain}	         &\textcolor{blue}{\underline{34.67}}	&32.68	&27.97 &24.98	&\textcolor{blue}{\underline{22.66}} \\[0.1em]
    AdaCoF\cite{lee2020adacof}	     &34.23	&32.16	&27.53	&24.84	&22.28 \\[0.1em]
    BMBC\cite{park2020bmbc} 	       &34.47	&32.13	&27.21	&24.52	&22.39 \\[0.1em]
    ABME (Proposed)	                     &\textcolor{red}{\bf 35.21}	&\textcolor{red}{\bf 34.15}	& \textcolor{red}{\bf 30.26}	& \textcolor{red}{\bf 25.77}	& \textcolor{red}{\bf 23.02} \\
    \bottomrule\\[-2em]
    \end{tabular}
    }
    \label{table:Evaluation on Xiph}
\end{table}

\begin{figure*}[t]
    \centering
	\subfloat {\includegraphics[width=1.7cm]{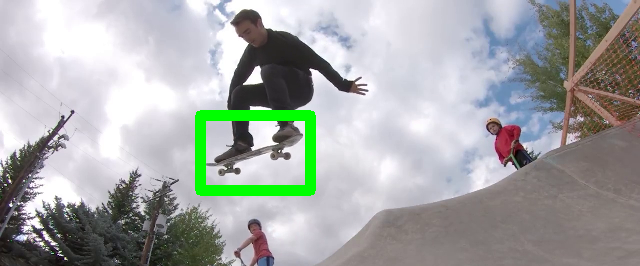}}\;\!\!
	\subfloat {\includegraphics[width=1.7cm]{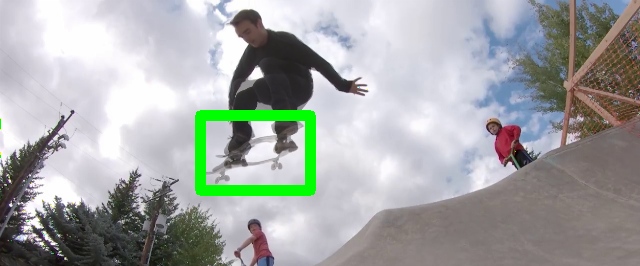}}\;\!\!
	\subfloat {\includegraphics[width=1.7cm]{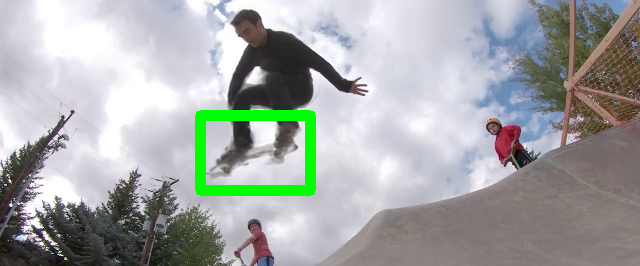}}\;\!\!
    \subfloat {\includegraphics[width=1.7cm]{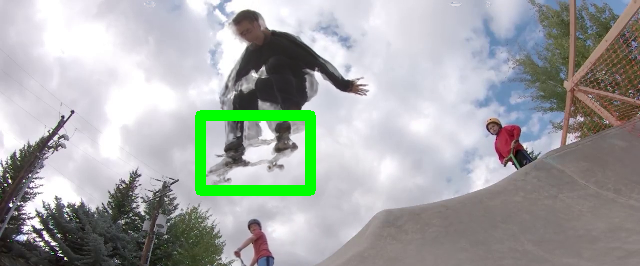}}\;\!\!
    \subfloat {\includegraphics[width=1.7cm]{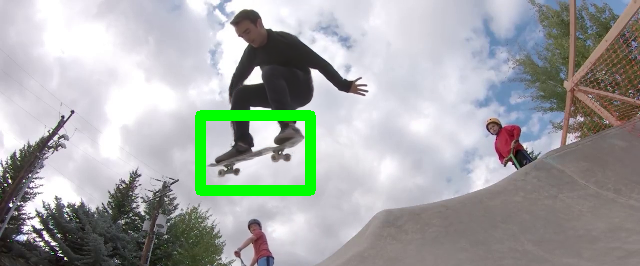}}\;\!\!
    \subfloat {\includegraphics[width=1.7cm]{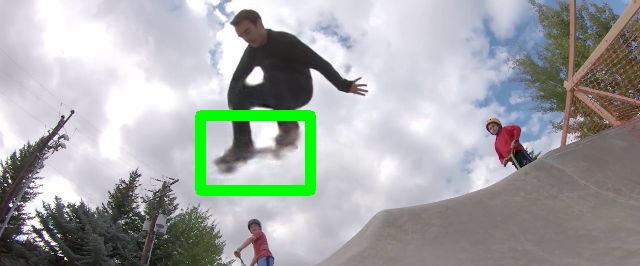}}\;\!\!
    \subfloat {\includegraphics[width=1.7cm]{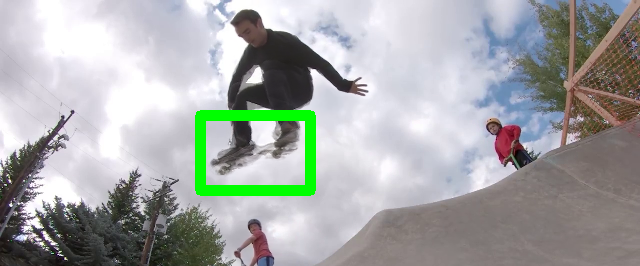}}\;\!\!
    \subfloat {\includegraphics[width=1.7cm]{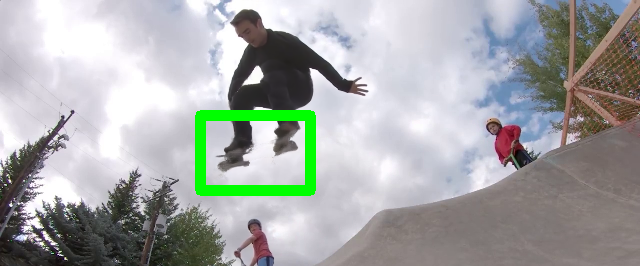}}\;\!\!
    \subfloat {\includegraphics[width=1.7cm]{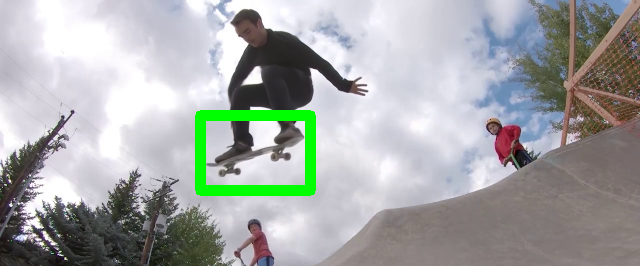}}\\[-1.0em]

    \subfloat {\includegraphics[width=1.7cm]{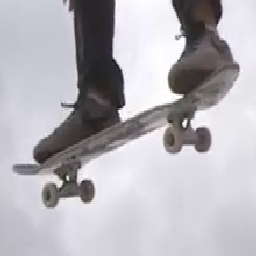}}\;\!\!
	\subfloat {\includegraphics[width=1.7cm]{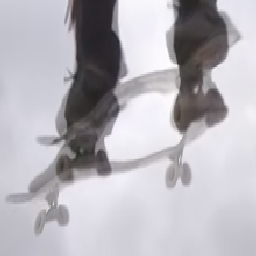}}\;\!\!
	\subfloat {\includegraphics[width=1.7cm]{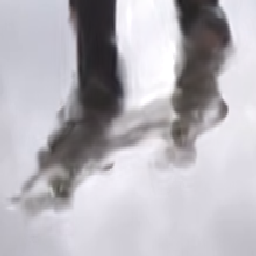}}\;\!\!
    \subfloat {\includegraphics[width=1.7cm]{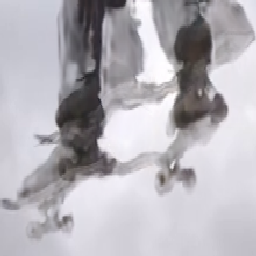}}\;\!\!
    \subfloat {\includegraphics[width=1.7cm]{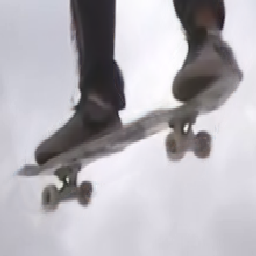}}\;\!\!
    \subfloat {\includegraphics[width=1.7cm]{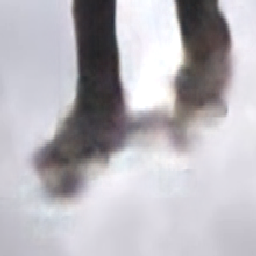}}\;\!\!
    \subfloat {\includegraphics[width=1.7cm]{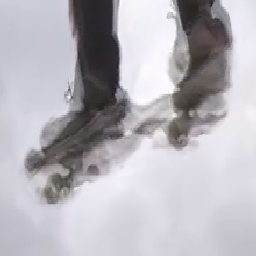}}\;\!\!
    \subfloat {\includegraphics[width=1.7cm]{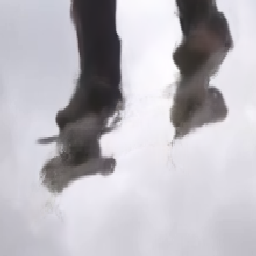}}\;\!\!
    \subfloat {\includegraphics[width=1.7cm]{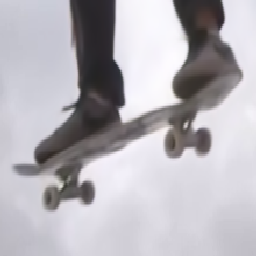}}\\[-0.9em]

    \setcounter{subfigure}{0}
    \subfloat {\includegraphics[width=1.7cm]{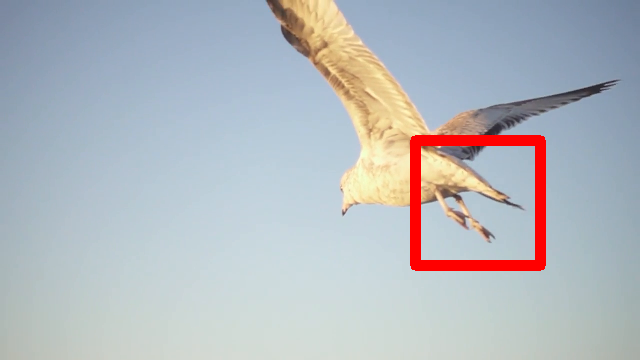}}\;\!\!
	\subfloat {\includegraphics[width=1.7cm]{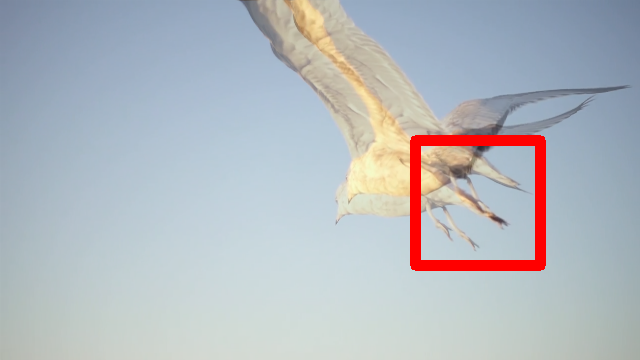}}\;\!\!
	\subfloat {\includegraphics[width=1.7cm]{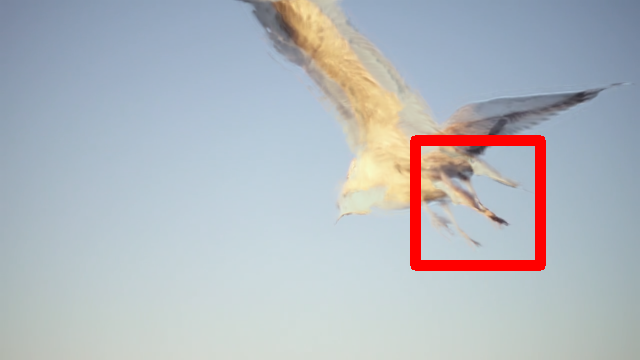}}\;\!\!
    \subfloat {\includegraphics[width=1.7cm]{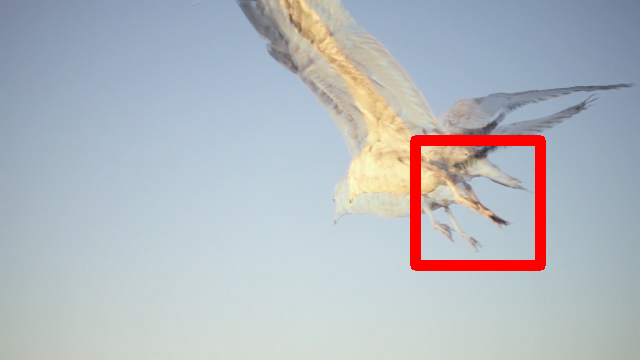}}\;\!\!
    \subfloat {\includegraphics[width=1.7cm]{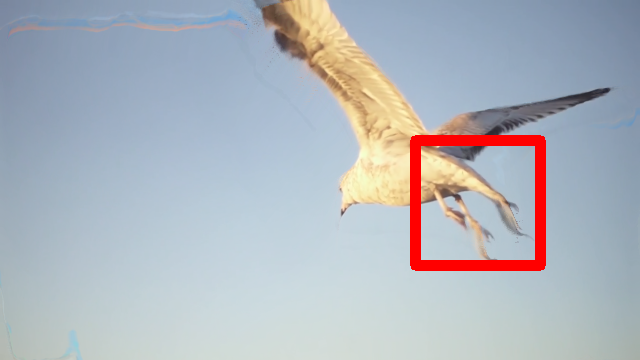}}\;\!\!
    \subfloat {\includegraphics[width=1.7cm]{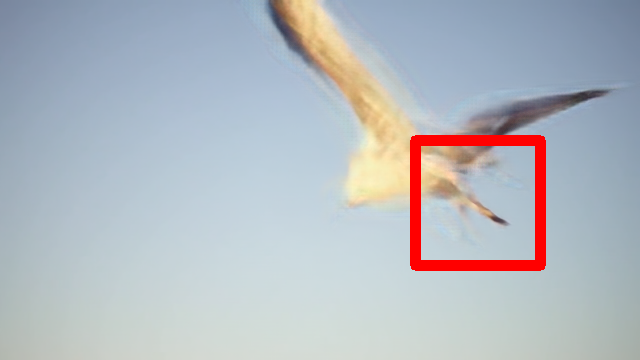}}\;\!\!
    \subfloat {\includegraphics[width=1.7cm]{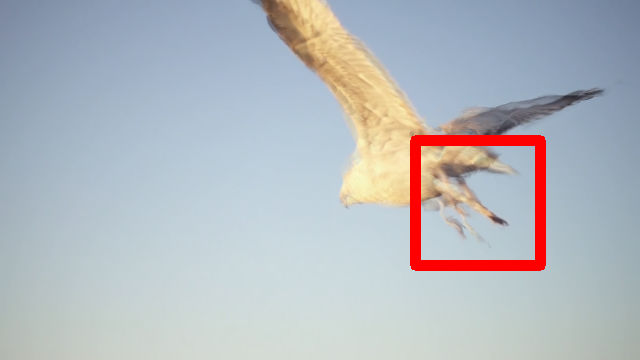}}\;\!\!
    \subfloat {\includegraphics[width=1.7cm]{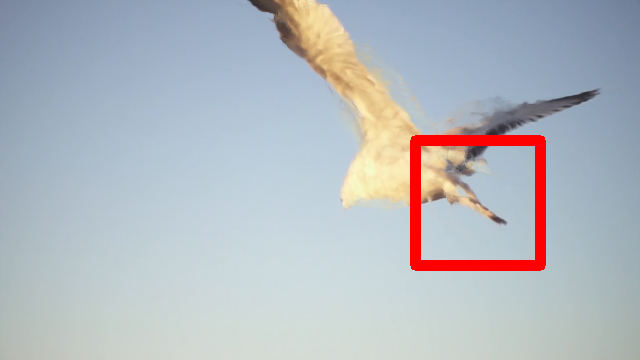}}\;\!\!
    \subfloat {\includegraphics[width=1.7cm]{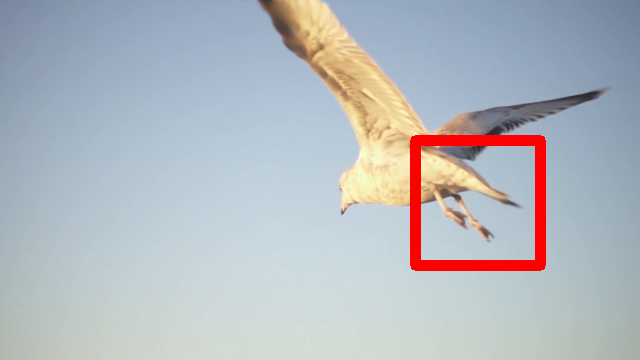}}\\[-1.0em]

    \setcounter{subfigure}{0}
	\subfloat [GT]{\includegraphics[width=1.7cm]{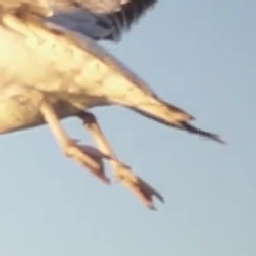}}\;\!\!
	\subfloat [ToFlow]{\includegraphics[width=1.7cm]{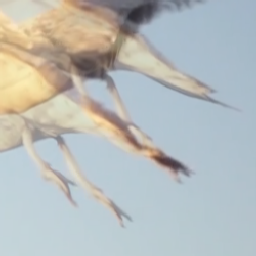}}\;\!\!
	\subfloat [SepConv]{\includegraphics[width=1.7cm]{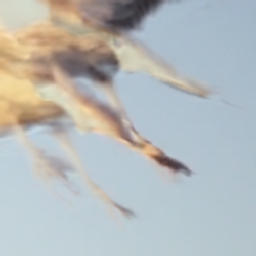}}\;\!\!
    \subfloat [CyclicGen]{\includegraphics[width=1.7cm]{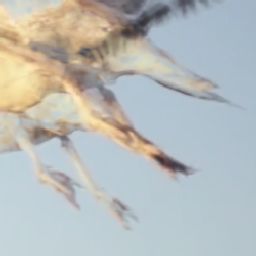}}\;\!\!
    \subfloat [DAIN]{\includegraphics[width=1.7cm]{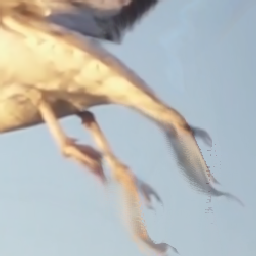}}\;\!\!
    \subfloat [CAIN]{\includegraphics[width=1.7cm]{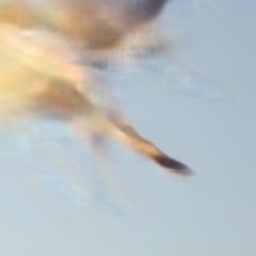}}\;\!\!
    \subfloat [AdaCoF]{\includegraphics[width=1.7cm]{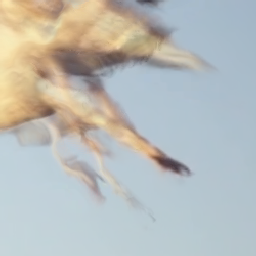}}\;\!\!
    \subfloat [BMBC]{\includegraphics[width=1.7cm]{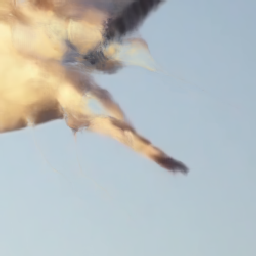}}\;\!\!
    \subfloat [ABME]{\includegraphics[width=1.7cm]{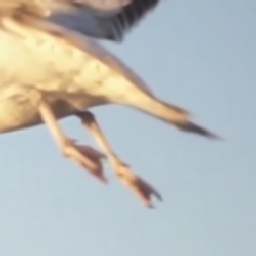}}\\
    \vspace*{0.05cm}
    \caption
	{
        Qualitative comparison of interpolated frames. Triplets in SNU-FILM (Extreme) are used in this test. The proposed ABME algorithm in (i) reconstructs rapid objects faithfully to the ground truth in (a) without noticeable artifacts.
	}
	\label{fig:Eval on SNU-FILM}
    \vspace*{-0.2cm}
\end{figure*}

\begin{figure}
  \vspace{-0.1cm}
  \centering
  \subfloat {\includegraphics[width=0.95\linewidth, height=5cm]{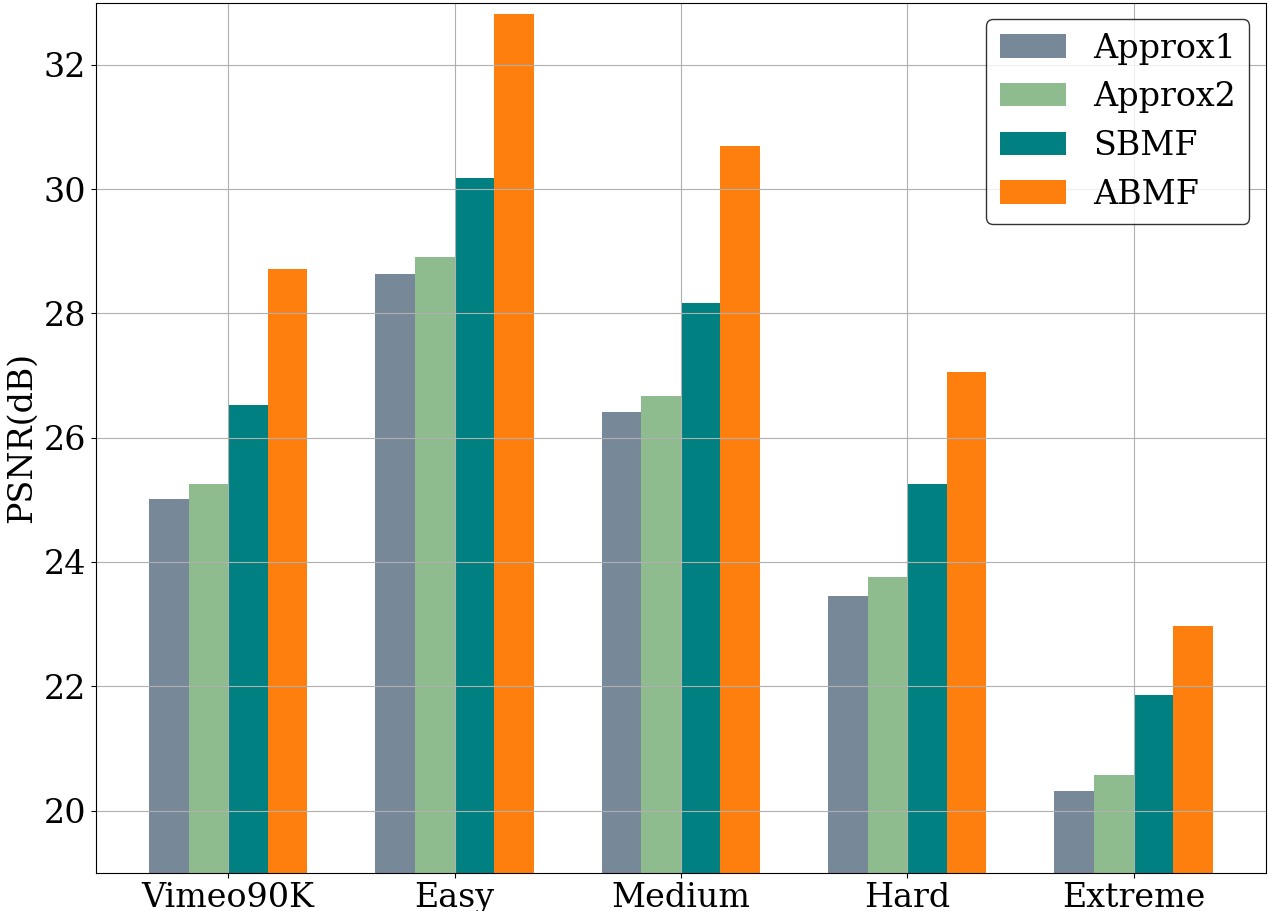}}
  \caption{Comparison of different motion fields.}
  \vspace{-0.2cm}
  \label{fig:intermediate_motion_graph}
\end{figure}

\subsection{Comparison with the State-of-the-Arts}

We compare the proposed algorithm with conventional algorithms: ToFlow \cite{xue2019toflow}, SepConv \cite{niklaus2017sepconv}, CyclicGen \cite{liu2019cyclicgen}, DAIN \cite{bao2019dain}, CAIN \cite{choi2020cain}, AdaCoF \cite{lee2020adacof}, and BMBC \cite{park2020bmbc}.

Table~\ref{table:Evaluation on test set} compares the average PSNR/SSIM scores. All results are obtained by executing available source codes.
\vspace*{-0.2cm}
\begin{itemize}
    \itemsep0mm
    \item UCF101, Vimeo90K, and FILM (Medium) are easier to interpolate than FILM (Hard, Extreme). Also, on the easiest FILM (Easy), most algorithms interpolate high quality frames, and visual differences between the results are negligible.
    \vspace*{-0.1cm}
    \item On all datasets except FILM (Easy), the proposed ABME algorithm provides the best results. Especially, on Vimeo90K, in comparison with the second best BMBC, ABME yields about 1 dB higher PSNR.
    \vspace*{-0.1cm}
    \item On FILM (Medium, Hard, Extreme), DAIN achieves the second best results in general; on FILM (Hard), ABME yields 30.58 dB, while DAIN 30.17 dB.
\end{itemize}
\vspace*{-0.2cm}
In videos with faster motions, more regions are occluded around motion boundaries, making it more difficult to interpolate frames. Table~\ref{table:Evaluation on test set} confirms that the proposed ABME handles these occluded regions effectively and provides excellent interpolation results.

Table~\ref{table:Evaluation on test set} also lists the actual runtime for interpolating an intermediate frame in the ``Urban" sequence in the Middlebury benchmark~\cite{baker2011database} using an RTX 2080 Ti GPU. The proposed algorithm is more than three times faster than BMBC.

Table~\ref{table:Evaluation on Xiph} compares the average PSNRs on the Xiph dataset, which contains diverse sequences with challenging factors, such as complicated texture, fast motion, and severe occlusion. ABME outperforms all conventional algorithms in all difficulty classes. DAIN achieves the second-highest PSNRs in classes D2, D3, and D4. However, its motion and depth estimators were pre-trained using additional datasets, whereas ABME was trained strictly using Vimeo90K only. Nonetheless, in D2 and D3, ABME outperforms DAIN by 0.94 dB and 0.84 dB, respectively. These results indicate that ABME interpolates challenging videos faithfully.

\begin{table}[t]
    \vspace*{0.1cm}
    \renewcommand{\arraystretch}{0.95}
    \caption
   {
        Quantitative comparison on the X4K1000FPS dataset.
   }
    \centering
    {\footnotesize
    \begin{tabular}{L{2.5cm}C{1.2cm}C{1.2cm}}
    \toprule
    & PSNR & SSIM \\
    \midrule
    DAIN\cite{bao2019dain}	         &26.78&0.8065\\[0.1em]
    AdaCoF\cite{lee2020adacof}	         &23.90&0.7271\\[0.1em]
    XVFI\cite{XVFI}	     &\textcolor{blue}{\underline{30.12}}&\textcolor{blue}{\underline{0.8704}}\\[0.1em]
    ABME (Proposed)	                 &\textcolor{red}{\bf 30.16}&\textcolor{red}{\bf 0.8793}\\
    \bottomrule\\[-2em]
    \end{tabular}
    }
    \label{table:Evaluation on X4K1000FPS}
\end{table}

Table~\ref{table:Evaluation on X4K1000FPS} compares the performances on the recent X4K1000FPS dataset. Its training set, X-TRAIN, is not used to retrain ABME. In contrast, X-TRAIN is used to train XVFI \cite{XVFI} that is designed for 4K sequences with extreme motions. Nevertheless, ABME performs slightly better than XVFI. It is a future research issue to tailor ABME for 4K sequences. For this purpose, the Vimeo90K training set might not be adequate, and X-TRAIN would be useful.

Figure~\ref{fig:Eval on SNU-FILM} shows interpolation results of two frames in FILM (Extreme). Because of large movements and extreme deformation, the conventional algorithms fail to reconstruct details within the green and red squares reliably. In contrast, the proposed algorithm reconstructs them faithfully without any noticeable artifacts.

\subsection{Model Analysis}
\label{ssec:modelA}

Let us analyze the contributions of key components in the proposed algorithm.

\vspace*{0.15cm}
{\noindent \bf Motion fields:} We compare the motion fields in Figure~\ref{fig:1D_motions} quantitatively. Specifically, we test the approximate motion field in (\ref{eq:m_approx_0}), called Approx1, and the approximate motion field in (\ref{eq:m_approx_2}), called Approx2, and the symmetric bilateral motion field (SBMF), and the proposed asymmetric bilateral motion field (ABMF). PWC-Net \cite{sun2018pwc} is used as the optical flow estimator for Approx1 and Approx2, while BMBC \cite{park2020bmbc} for SBMF. These four types of motion fields are used, respectively, to warp input frames backward to approximate the intermediate frame, and then the PSNRs of the warped frames are computed. Figure~\ref{fig:intermediate_motion_graph} compares the average PSNRs on Vimeo90K and SNU-FILM. Approx1 produces the worst performance. Approx2, which improves the approximation via (\ref{eq:m_approx_2}), slightly increases the PSNRs. SBMF in \cite{park2020bmbc} yields considerably higher PSNRs than Approx1 or Approx2. However, the proposed ABMF outperforms SBMF significantly by more than 2 dB on Vimeo90K, Easy, and Medium and at least 1 dB on Hard and Extreme.

\begin{figure}
  \vspace{-0.4cm}
  \centering
  \subfloat {\includegraphics[width=1.9cm]{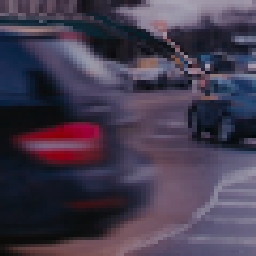}}\!
  \subfloat {\includegraphics[width=1.9cm]{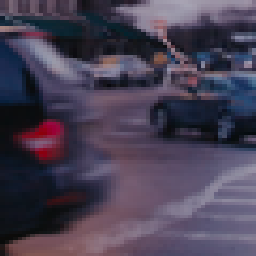}}\!
  \subfloat {\includegraphics[width=1.9cm]{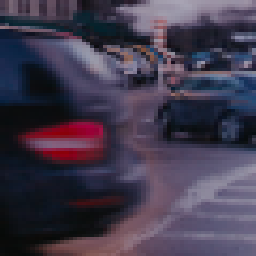}}\!
  \subfloat {\includegraphics[width=1.9cm]{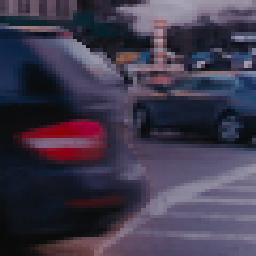}}\!
  \subfloat {\includegraphics[width=0.565cm]{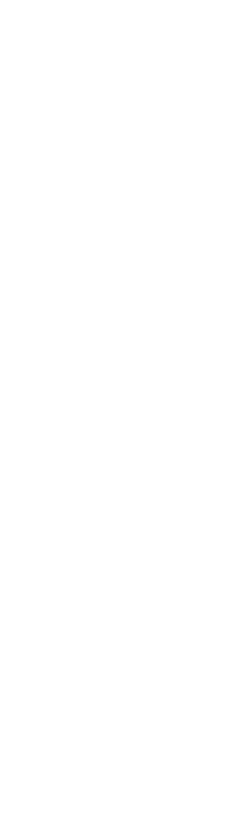}}\\[-1.0em]

  \setcounter{subfigure}{0}
  \subfloat [Approx1]{\includegraphics[width=1.9cm]{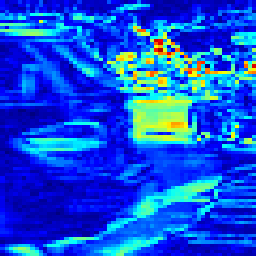}}\!
  \subfloat [Approx2]{\includegraphics[width=1.9cm]{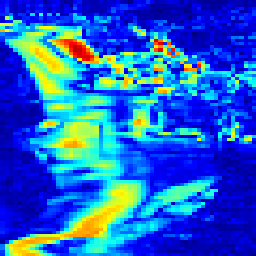}}\!
  \subfloat [SBMF]{\includegraphics[width=1.9cm]{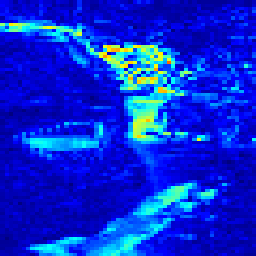}}\!
  \subfloat [ABMF]{\includegraphics[width=1.9cm]{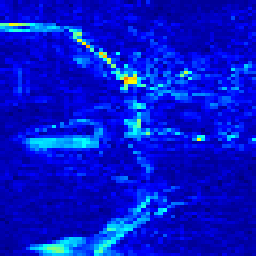}}\!
  \subfloat {\includegraphics[width=0.565cm]{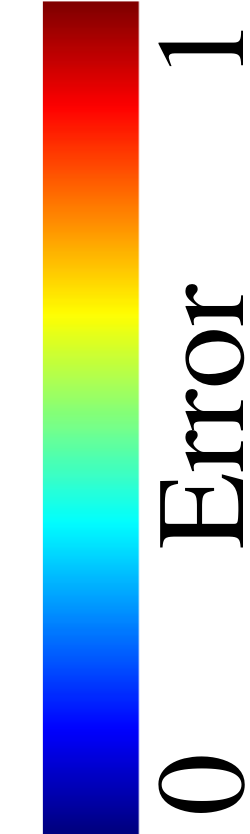}}
  \vspace*{0.05cm}
  \caption{Comparison of warped frames and their error maps.}
  \vspace{-0.2cm}
  \label{fig:intermediate_motion_estimation}
\end{figure}

Figure~\ref{fig:intermediate_motion_estimation} compares warped frames with error maps.
The car moves quickly, causing severe occlusion. Hence, Approx1 and Approx2 distort the warped frames severely. SBMF estimates the motion of the car accurately but causes errors around the boundary of the car due to disocclusion. In contrast, the proposed ABMF reduces these errors effectively by employing asymmetric motion vectors.

\vspace*{0.15cm}
{\noindent \bf Reliability mask and offset map:}
We analyze the effectiveness of ABMR-Net in Figure~\ref{fig:motion_networks}. It uses a learnable reliability mask $Z_t$ and a learnable offset map $O_t$. Table~\ref{table:modified_pwc_net} shows that both components contribute to the overall performance of the asymmetric bilateral motion refinement.

\begin{table}[t]
    \renewcommand{\arraystretch}{0.95}
    \caption
    {
        Impacts of the reliability mask $Z_t$ and the offset map $O_t$ on the asymmetric bilateral motion refinement. Average PSNRs of warped frames are reported.
    }
    \small
    \centering
    \begin{tabular}{C{1.5cm}C{1.5cm}C{1.5cm}C{1.5cm}}
    \toprule
    \multicolumn{2}{c}{Component}& Vimeo90K & Extreme\\
    \cmidrule(lr){1-2} \cmidrule(lr){3-3} \cmidrule(lr){4-4}
     $Z_t$ & $O_t$ & PSNR & PSNR\\[-0.1em]
    \midrule
      &  & 27.81 & 22.75\\
    \checkmark &  &28.32  &22.96 \\
    &\checkmark  &28.40  &22.98 \\
    \checkmark & \checkmark & 28.80 & 23.07\\[-0.1em]
    \bottomrule\\[-1em]
    \end{tabular}
    \label{table:modified_pwc_net}
\end{table}

\begin{table}[t]
    \renewcommand{\arraystretch}{0.95}
    \caption
    {
        Impacts of candidate warped frames on the frame synthesis. Average PSNRs of interpolated frames are reported.
    }
    \small
    \centering
    \begin{tabular}{C{3.3cm}C{1.5cm}C{1.5cm}}
    \toprule
    \multirow{2}[1]{*}{Candidates}& Vimeo90K & Extreme\\
    \cmidrule(lr){2-2} \cmidrule(lr){3-3}
    & PSNR & PSNR\\[-0.1em]
    \midrule
    Symmetric only & 35.91 & 25.28\\
    Asymmetric only & 36.05 & 25.34\\
    Both &36.18 & 25.42\\[-0.1em]
    \bottomrule
    \end{tabular}
    \label{table:Intermeidate candidates}
    \vspace*{0cm}
\end{table}

\vspace*{0.15cm}
{\noindent \bf Candidate warped frames:} In the proposed frame synthesis in Figure~\ref{fig:overview_synthesis}, four candidate warped frames are used in the default mode. Specifically, two candidates are obtained using the symmetric fields ${\cal V}^{\textrm{S}}_{t \rightarrow 0}$ and ${\cal V}^{\textrm{S}}_{t \rightarrow 1}$, and the other two from the asymmetric fields ${\cal V}^{\textrm{A}}_{t \rightarrow 0}$ and ${\cal V}^{\textrm{A}}_{t \rightarrow 1}$. We test the following three combinations.
\vspace*{-0.2cm}
\begin{itemize}
    \itemsep0mm
    \item Symmetric only: Two candidates using the symmetric fields are used.
    \vspace*{-0.1cm}
    \item Asymmetric only: Two candidates using the asymmetric fields are used.
    \vspace*{-0.1cm}
    \item Both: All four candidates are used.
\end{itemize}
\vspace*{-0.2cm}
Table~\ref{table:Intermeidate candidates} compares the average PSNRs of these three settings. First, `Symmetric only' yields the worst performance, while it exceeds the performance of BMBC~\cite{park2020bmbc}. This means that the proposed synthesis network is more effective than that of BMBC. Second, `Asymmetric only'
performs better than `Symmetric only' by loosening the linear motion constraint. Last, the best PSNRs are achieved by employing all four candidates, which complement each other to improve the interpolation performance.

\begin{figure}
  \vspace{-0.2cm}
  \centering
  \subfloat {\includegraphics[width=\linewidth]{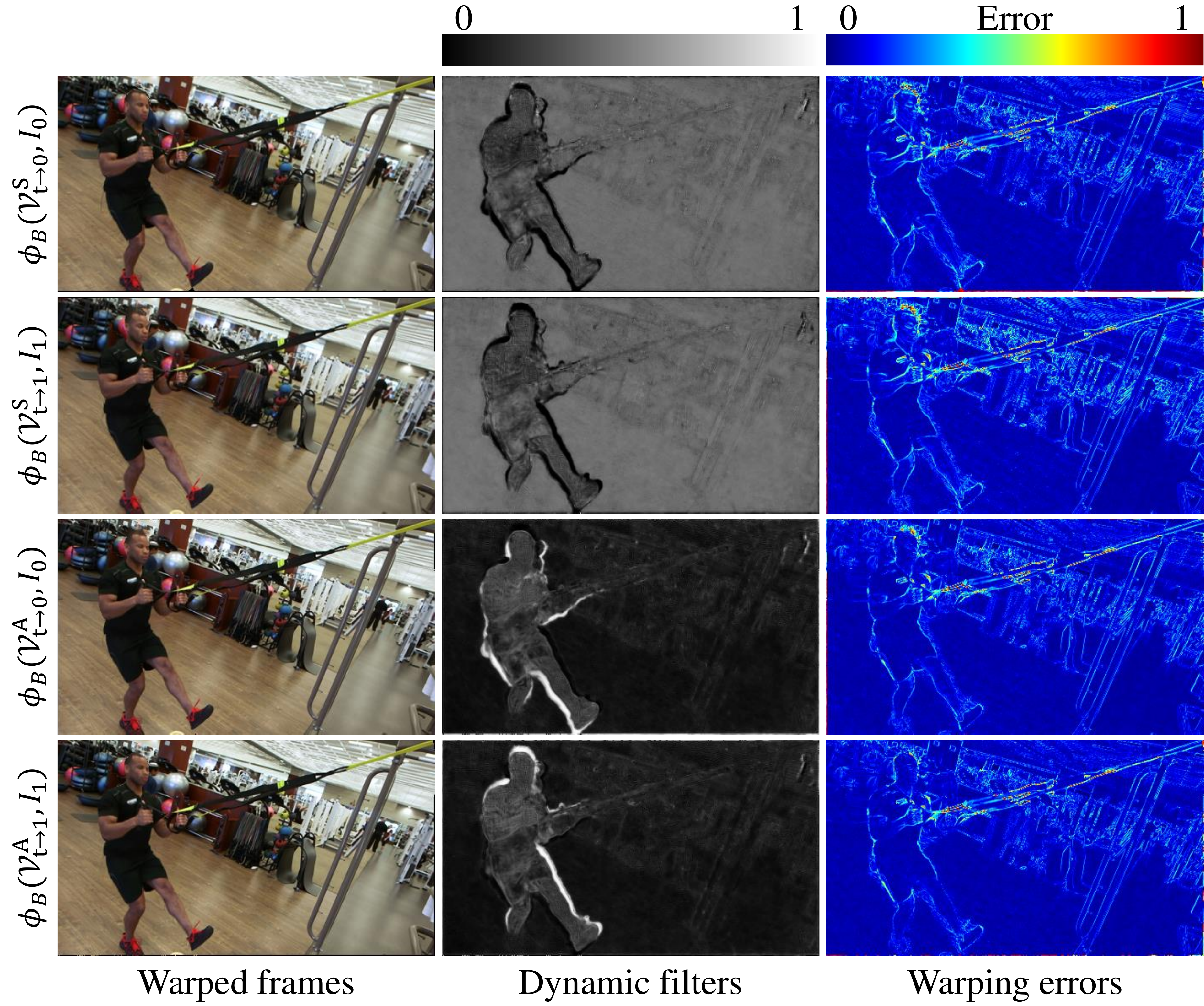}}\\
  \caption{Visualization of dynamic filters.}\label{fig:dynamic_filter}
  \vspace{-0.2cm}
\end{figure}

{\noindent \bf Dynamic filters:}
Figure~\ref{fig:dynamic_filter} shows the four candidate warped frames, their filter coefficient maps, and their error maps. To visualize dynamic filters, we compute the absolute sum of coefficients for each pixel and render the sum in a gray level. The top two candidates are obtained by the symmetric fields, while the others by the asymmetric ones. The former are more reliable in the background without motion, while the latter are more effective around motion boundaries. Hence, the filter coefficients are determined accordingly. Moreover, to the left and right sides of the man, the third and fourth candidates are mainly used for the dynamic filtering, respectively. This is because different sides are occluded in different frames $I_0$ and $I_1$. In such occluded regions, the linear motion constraint is invalid and the symmetric fields are not reliable. 

\vspace*{0.15cm}
{\noindent \bf Filtered feature maps:} In Figure~\ref{fig:overview_synthesis}, RefineNet takes the filtered feature map $\widetilde{C}_t$, as well as the filtered frame $\tilde{I}_t$, as input. If $\widetilde{C}_t$ is removed and only $\tilde{I}_t$ is used to synthesize $I_t$, the average PSNR is reduced by 0.1 dB on Vimeo90K. Since $\widetilde{C}_t$ conveys contextual information in $\tilde{I}_t$, it helps RefineNet to remove noise in $\tilde{I}_t$ and restore $I_t$ more faithfully.

\section{Conclusions}
We proposed a novel, effective video frame interpolation algorithm. First, we developed the ABME technique to refine symmetric bilateral motions by loosening the linear motion constraint. Second, we designed the novel synthesis network that generates a set of dynamic filters and a residual frame using local and global information. Extensive experiments demonstrated that the proposed algorithm achieves state-of-the-art performance on various datasets.

\section*{Acknowledgements}
\noindent This work was supported by the National Research Foundation of Korea (NRF) grants funded by the Korea government (MSIT) (Nos.~NRF-2018R1A2B3003896, NRF-2019R1A2C4069806, and NRF-2021R1A4A1031864).

{\small
\bibliographystyle{ieee_fullname}
\bibliography{egbib}
}

\end{document}